%% file: main.tex
\documentclass[10pt,twocolumn,letterpaper]{article}
\usepackage[pagenumbers]{wacv}

\input{preamble}

\definecolor{wacvblue}{rgb}{0.21,0.49,0.74}
\usepackage[pagebackref,breaklinks,colorlinks,allcolors=wacvblue]{hyperref}

\title{PlenoCI: Plenoptic CharacterIstics for\\ View Dependence Aware Change Classification}

\author{Jason Lai$^1$ \hspace{10pt} Chamuditha Jayanga Galappaththige$^{2,3}$ \hspace{10pt} \\
Niko S\"underhauf$^{2,3}$ \hspace{10pt}  Dimity Miller$^{2,3}$ \hspace{10pt} Donald G. Dansereau$^{1,3}$ \\
$^1$Australian Centre for Robotics, School of Aerospace, Mechanical and Mechatronic Engineering,\\The University of Sydney \hspace{1pt} $^2$QUT Centre for Robotics \hspace{1pt} $^3$ARIAM Hub\\
{\tt\small \{jason.lai, donald.dansereau\}@sydney.edu.au}
}

\begin{document}

\maketitle
\input{0_abstract}    
\glsresetall
\input{1_intro}
\input{2_related}
\input{3_plenoci}
\input{4_changes}
\input{5_results}
\input{6_conclusion}
\input{7_acknowledgements}
{
    \small
    \bibliographystyle{ieeenat_fullname}
    \bibliography{main}
}

\input{supplementary}

\end{document}

%% file: preamble.tex
\usepackage{multirow}
\usepackage{mathrsfs} 
\usepackage{stfloats}
\usepackage[nolist]{glossaries}
\glsdisablehyper 
\input{glossary}

\DeclareMathOperator*{\argmax}{arg\,max}

\definecolor{goldbg}{RGB}{255, 215, 0}
\definecolor{silverbg}{RGB}{230, 230, 230}   
\definecolor{bronzebg}{RGB}{218, 165, 32}

\newcommand{\first}[1]{\colorbox{goldbg}{#1}}
\newcommand{\second}[1]{\colorbox{silverbg}{#1}}
\newcommand{\third}[1]{\colorbox{bronzebg}{#1}}



%% file: glossary.tex
\newacronym{NeRF}{NeRF}{neural radiance fields}
\newacronym{3DGS}{3DGS}{3D Gaussian Splatting}
\newacronym{SH}{SH}{spherical harmonics}
\newacronym{SfM}{SfM}{structure from motion}
\newacronym{mIoU}{mIoU}{mean intersection over union}
\newacronym{SCD}{SCD}{scene change detection}
\newacronym{CC}{CC}{change classification}
\newacronym{PlenoCI}{PlenoCI}{Plenoptic CharacterIstics}
\newacronym{FLOPs}{FLOPs}{floating point operations}
\newacronym{FPR}{FPR}{false positive rate}

%% file: 0_abstract.tex
\begin{abstract}
Radiance field representations such as \gls{3DGS} natively encode complex visual phenomena such as occlusions and view dependence, but they are inherently underconstrained.
Independently optimized reconstructions converge to different primitive configurations, even in unchanged regions.
We introduce \gls{PlenoCI}, a novel feature built from the plenoptic field these representations approximate.
\gls{PlenoCI} directly captures rich visual behaviors while ignoring Lambertian textures.
By deriving closed-form analytic plenoptic derivatives from a \gls{3DGS} representation, we efficiently detect these 5D structures.
Our approach is robust to underconstrained representations by construction, reporting two orders of magnitude fewer false positives between independent reconstructions of unchanged scenes than concurrent work.
We demonstrate \gls{PlenoCI}'s utility on change classification.
First, we detect changes with an instance-aware \gls{3DGS} pipeline, achieving state-of-the-art results on CL-Splats with a 25.7\% mIoU gain over the strongest competitor, while remaining competitive on the more challenging PASLCD benchmark.
Leveraging \gls{PlenoCI}, we classify changes as geometric or appearance-based with a balanced accuracy of 0.735, comparable to the best performing baseline.
We believe plenoptic derivatives and \gls{PlenoCI} open new directions for view dependence aware understanding in visually complex environments.
Code and data are available at \href{https://js0n-lai.github.io/plenoci}{https://js0n-lai.github.io/plenoci}.
\end{abstract}

%% file: 1_intro.tex
\section{Introduction} \label{sec:intro}

\begin{figure*}[tb]
    \centering
    \includegraphics[width=\linewidth]{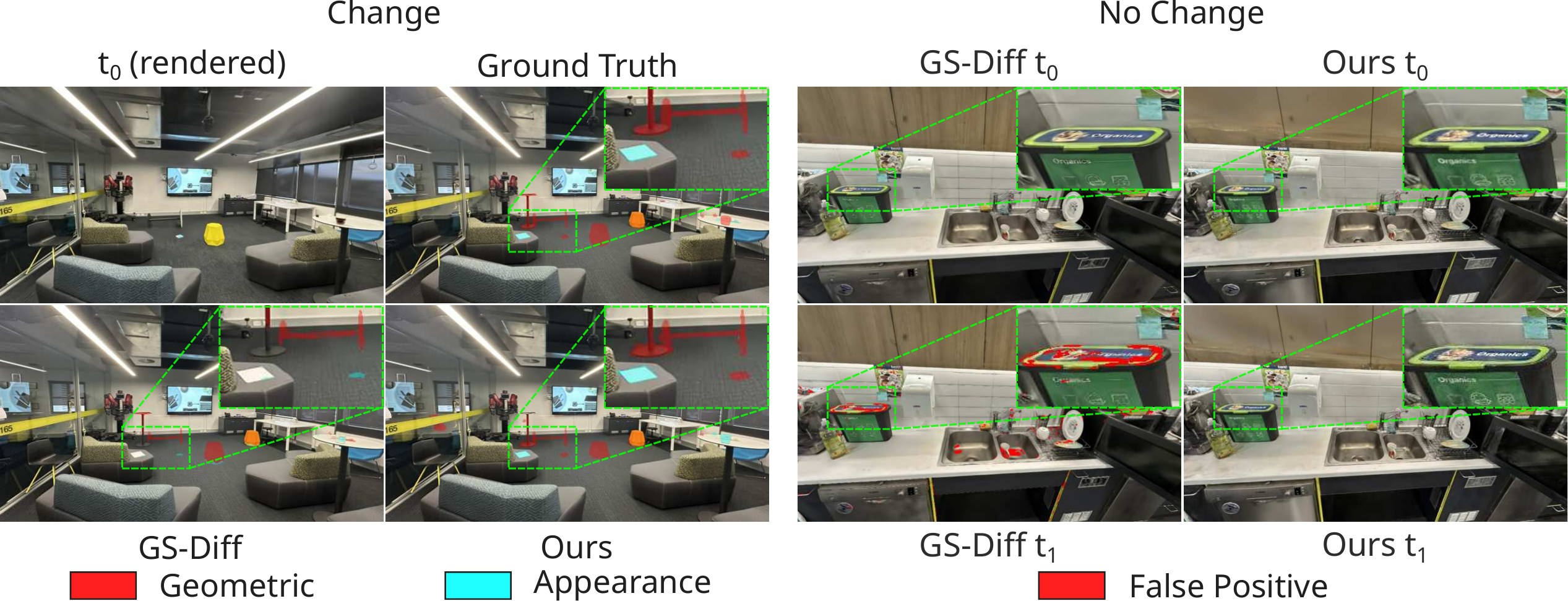}
    \caption{Given two \acrshort{3DGS} reconstructions from different times, our method first detects scene changes. To classify changes, we draw on view-dependent plenoptic structures encoded in our novel \gls{PlenoCI} feature. For example, we correctly detect the sheet of paper and book as changes, and classify them using \gls{PlenoCI} as occlusion edge evidence (left). Our approach is robust to the inherently underconstrained nature of \acrshort{3DGS} representations; two independent optimizations given the same training images yield different primitive configurations. Primitive-space comparisons~\cite{galappaththige2026gsdiff} report false positives in static scenes while our method does not (right).}
    \label{fig:teaser}
\end{figure*}

Robust \gls{SCD} and \gls{CC} are vital for autonomous systems in unconstrained environments, with applications spanning damage assessment~\cite{sakurada_change_2015}, environmental monitoring~\cite{taneja2011image}, and map updating~\cite{radke2005image}.
For robotic platforms tasked with long-term monitoring, such as routine operator rounds in industrial facilities, reliably distinguishing genuine geometric and appearance changes from superficial lighting fluctuations, as our approach does in \Cref{fig:teaser}, enables richer scene understanding and drives advanced autonomous reasoning.

Radiance field representations such as \gls{NeRF}~\cite{mildenhallNeRFRepresentingScenes2020} and \gls{3DGS}~\cite{kerbl3DGaussianSplatting2023} natively capture complex visual phenomena such as occlusion edges, specularity and reflections common in unconstrained settings.
However, recent \gls{SCD} works~\cite{lu3DGSCD3DGaussian2025,Galappaththige_2025_CVPR,galappaththige2025changes,galappaththige2026gsdiff,zhou20253d} building on these representations treat view-dependent effects purely as inconsistencies to suppress.

Our approach uses three insights: view dependence can signal complex visual behaviors; these phenomena manifest as unique structures in the 5D plenoptic field, captured by our novel \acrfull{PlenoCI} feature;
and regressing the plenoptic field with \gls{3DGS} yields closed-form, analytic derivatives that make these structures efficient to detect.
Notably, \gls{PlenoCI} extraction requires no per-scene retuning across all scenes and datasets we evaluate on.
By construction, \gls{PlenoCI} also bypasses a fundamental challenge: radiance field representations are underconstrained.
As \Cref{fig:teaser} shows, two independent optimizations given the same training images converge to different primitive configurations.
Concurrent work~\cite{galappaththige2026gsdiff} addresses this by explicitly modeling reconstruction drift to enable primitive comparisons.
We take an orthogonal approach, deriving \gls{PlenoCI} from the plenoptic field these representations approximate.
We report a false positive rate two orders of magnitude lower when comparing independently optimized \gls{3DGS} models of unchanged scenes.

We demonstrate these tools on offline \gls{CC}, where captures follow unconstrained trajectories through a scene.
We detect changes with an instance-aware \gls{3DGS} pipeline that aggregates change cues across views, then use \gls{PlenoCI} as a proxy for scene geometry to distinguish geometric from appearance-based changes.

Our key contributions are:
\begin{itemize}
    \item We derive closed-form analytic plenoptic derivatives from a \gls{3DGS} representation, and show they can be rendered efficiently using standard \gls{3DGS} rasterization.
    \item We propose \gls{PlenoCI} as a novel 5D feature for summarizing informative plenoptic structures manifesting as complex visual behaviors, while ignoring Lambertian textures. These features build on generalized structure tensor analysis enabled by our plenoptic derivatives.
    \item We introduce a \gls{3DGS}-based \gls{SCD} pipeline using off-the-shelf instance segmentation to detect changed objects, and combine this with \gls{PlenoCI} to distinguish between geometric and appearance-based changes.
\end{itemize}
We believe this work advances robust scene understanding.
\gls{PlenoCI} has possible applications in tasks such as localization, navigation and map stitching, while plenoptic derivatives can be used for visual odometry and uncertainty estimation.
Code will be released upon acceptance.

%% file: 2_related.tex
\section{Related Work} \label{sec:related}

\subsection{2D Scene Change Detection}

Early \gls{SCD} works compare 2D images, relying on fully or weakly supervised training on closely aligned image pairs~\cite{sakuradaWeaklySupervisedSilhouettebased2020,caye_daudt_fully_2018,vargheseChangeNetDeepLearning2018,wangHowReduceChange2023,sachdeva2023change,alcantarilla2018street,chen2021dr,lei2020hierarchical,varghese2018changenet,lee2024semi,linRobustSceneChange2024,sachdevaChangeYouWant2023}. However, this paradigm is significantly hindered by costly change annotations and real-world domain shifts~\cite{gulrajani2021in}. 
Vision foundation models~\cite{oquab2024dinov2,kirillovSegmentAnything2023} have introduced a new paradigm for zero-shot \gls{SCD}~\cite{Kim_2025_CVPR,cho2025zero,kannan_2025,alpherts2025emplace}, circumventing the drawbacks of supervised models and outperforming them on real-world datasets. These approaches however inherit the failure modes of their underlying backbones while remaining brittle against viewpoint discrepancies and view-dependent inconsistencies~\cite{galappaththige2025changes}. In contrast, our approach aggregates change cues across multiple views, is agnostic to viewpoint variations, and can additionally classify changes.

\subsection{2D-3D Scene Change Detection}

Rich 3D representations can be constructed from multiview images for \gls{SCD}.
Geometric foundation models~\cite{wang2025vggt,mast3r,wang2026pi} directly infer 3D scene properties using attention mechanisms.
However, \gls{SCD} works~\cite{friedlander2026goldilocs,liu2025leveraging,wu2026scenediff} operating on these representations inherently struggle in scenes with strongly view-dependent behavior.
\gls{NeRF}~\cite{mildenhallNeRFRepresentingScenes2020} and \gls{3DGS}~\cite{kerbl3DGaussianSplatting2023} have been widely established as high-fidelity, photorealistic representations capable of capturing view dependence. Consequently, recent methods in \gls{SCD} use \gls{NeRF}~\cite{huangCNERFRepresentingScene2023,zhouPADDatasetBenchmark2023,martinsonMeaningfulChangeDetection2024} and \gls{3DGS}~\cite{kruseSplatPoseDetectPoseAgnostic,liuSplatPoseRealtimeImageBased2024,Galappaththige_2025_CVPR,lu3DGSCD3DGaussian2025,jiangGaussianDifferenceFind2025,galappaththige2025changes,zhou20253d,hattori_2026} to model the pre-change scene and synthesize views from the post-change viewpoint, enabling \gls{SCD} in a pose-agnostic setting.
These methods use learned semantic features~\cite{kruseSplatPoseDetectPoseAgnostic,liuSplatPoseRealtimeImageBased2024,lu3DGSCD3DGaussian2025,zhou20253d,hattori_2026} from vision backbones, or a combination with pixel data~\cite{galappaththige2025changes,Galappaththige_2025_CVPR} to generate zero-shot change cues.

Recent works~\cite{galappaththige2025changes,Galappaththige_2025_CVPR,lu3DGSCD3DGaussian2025,zhou20253d,galappaththige2026predictive} move beyond simple pairwise comparisons, aggregating change information across views to achieve state-of-the-art results.
However, these approaches treat view-dependent cues purely as inconsistencies to suppress, discarding possibly genuine changes.
Concurrent work~\cite{galappaththige2026gsdiff} detects changes in primitive space with explicit drift modeling to overcome underconstrained optimizations, but discards higher order \gls{SH} encoding view dependence.
We take an orthogonal approach, explicitly uncovering view-dependent structures by operating in the plenoptic field.

\subsection{Leveraging the Plenoptic Field}

The plenoptic field~\cite{adelsonPlenopticFunctionElements1991} describes light rays as a high-dimensional phenomenon, mapping position, direction, wavelength and time to intensity.
Light fields~\cite{levoyLightFieldRendering1996,szeliskiLumigraph1996} represent 4D slices with fixed wavelength and time, while assuming rays are constant-valued along their propagation.
They are applicable in tasks such as depth estimation~\cite{dansereau_depth_2004}, \gls{SfM}~\cite{tsai_2019,johannsen_2015}, scene flow~\cite{Ma_2018_ECCV} and visual odometry~\cite{neumann_2002,dansereau2011plenoptic}.
Dansereau \etal~\cite{dansereauSimpleChangeDetection2016} propose closed-form change detection as regions that violate plenoptic flow, a linear system relating light field derivatives to camera motion.
However, these methods derive analytic derivatives assuming a Lambertian scene, or densely sample for numerical estimation.
By leveraging \gls{3DGS} as a plenoptic regression, we derive closed-form analytic expressions with no such assumptions, and generalize beyond light fields.

Radiance field representations including \gls{NeRF}~\cite{mildenhallNeRFRepresentingScenes2020} and \gls{3DGS}~\cite{kerbl3DGaussianSplatting2023} regress the 5D plenoptic function~\cite{fridovich_2022_plenoxels,freitasComparativeAssessmentImplicit2024,Galappaththige_2025_CVPR,petrovska_2025,freitas_2026}.
Recent works use the plenoptic field as a unifying space, such as benchmarking representations~\cite{freitasComparativeAssessmentImplicit2024}, and geometry regularization with surface light field constraints~\cite{naylor2025surf}. 
Building on this perspective, we efficiently detect 5D plenoptic structures from a \gls{3DGS} representation.

%% file: 3_plenoci.tex
\section{PlenoCI: A New Plenoptic Feature} \label{sec:plenoci}

We introduce \gls{PlenoCI}, a feature capturing rays that experience color change as they shift along multiple axes of position and direction, signaling complex behaviors such as occlusion edges and specularity.
We build intuition in a 2D world (\cref{sec:flatland}) before generalizing to 3D scenes (\cref{sec:plenoci_generalizing}).

\subsection{Toy Example: Flatland} \label{sec:flatland}

\begin{figure}[tb]
    \centering
    \begin{subfigure}{0.48\linewidth}
        \includegraphics[width=\linewidth]{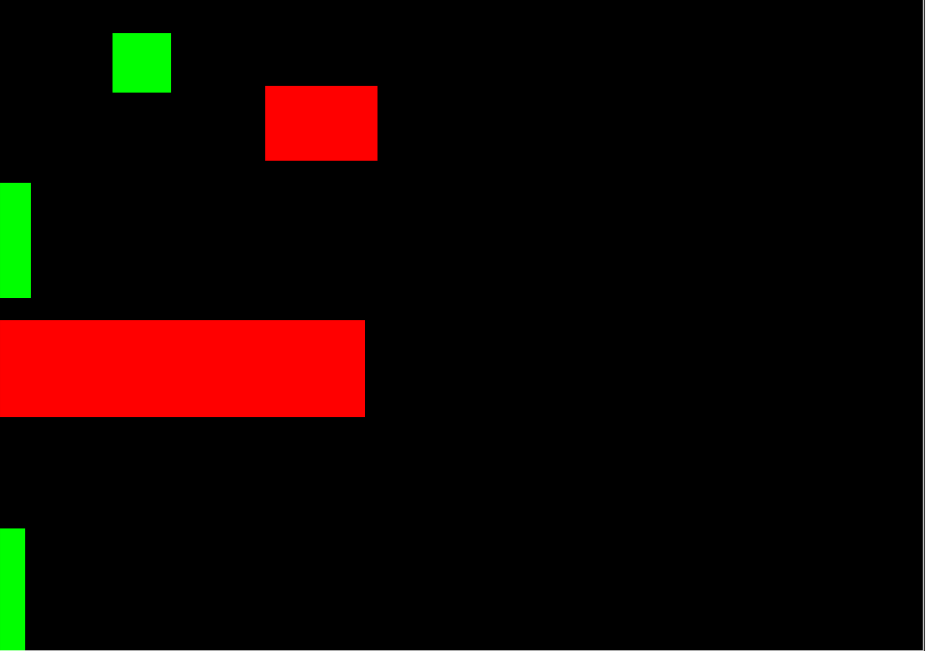}
        \caption{Example scene}
        \label{fig:flatland_env}
    \end{subfigure}
    \hfill
    \begin{subfigure}{0.48\linewidth}
        \includegraphics[width=\linewidth]{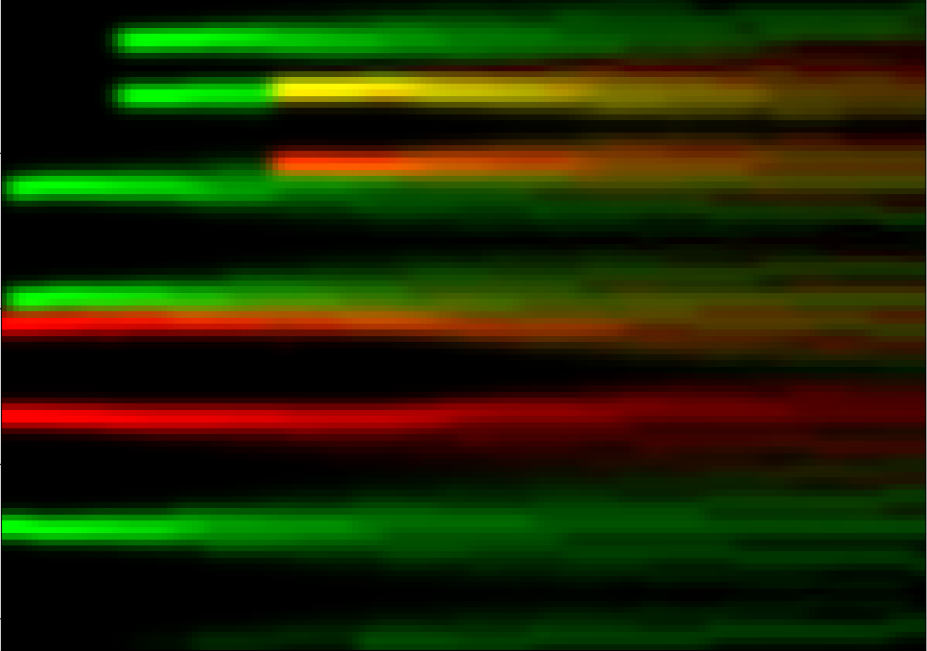}
        \caption{$\lvert\mathscr{L}_y\rvert$ at $\theta=180^{\circ}$}
        \label{fig:flatland_dy}
    \end{subfigure}
    \begin{subfigure}{0.48\linewidth}
        \includegraphics[width=\linewidth]{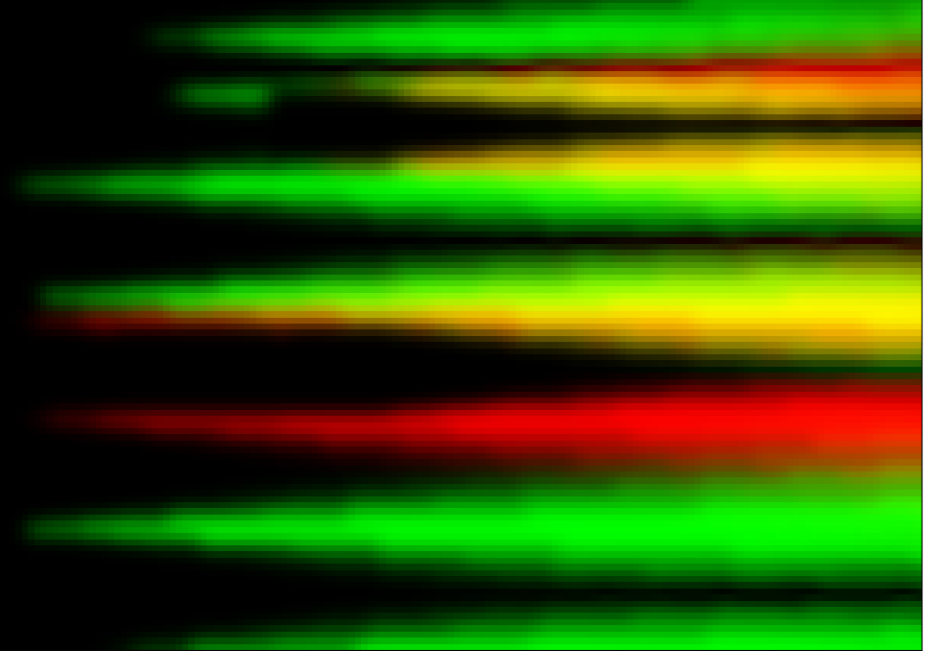}
        \caption{$\lvert\mathscr{L}_{\theta}\rvert$ at $\theta=180^{\circ}$}
        \label{fig:flatland_dt}
    \end{subfigure}
    \hfill
    \begin{subfigure}{0.48\linewidth}
        \includegraphics[width=\linewidth]{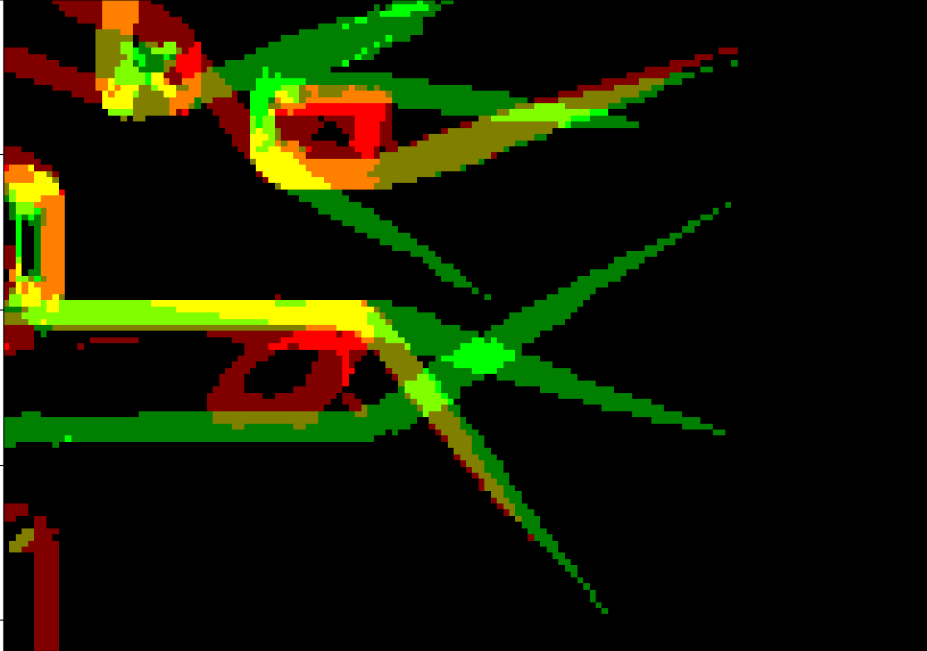}
        \caption{Occlusion edges}
        \label{fig:flatland_occlusion_crossings}
    \end{subfigure}
    \caption{Example scenario in flatland to motivate our novel \gls{PlenoCI} feature. Given a 2D world (\cref{fig:flatland_env}), we build the 3D plenoptic field $\mathscr{L}$ and compute partial derivatives $\mathscr{L}_{*}$ (\crefrange{fig:flatland_dy}{fig:flatland_dt}). These are useful signals for detecting visually complex phenomena such as occlusion edges (\cref{fig:flatland_occlusion_crossings}). Brightness represents magnitude corresponding to each color channel's plenoptic field. While we show ground truth geometry and $\mathscr{L}$ here, this also generalizes to radiance field representations.}
  \label{fig:flatland_setup}
\end{figure}

We begin exploring the plenoptic field $\mathscr{L}(\mathbf{r})=\mathscr{L}(x,y,\theta)$ in flatland, a world with two spatial dimensions $(x,y)$ and in-plane rotation $\theta$.
\Cref{fig:flatland_env} shows an example scene.
The problem is stated as follows: a set of 1D images $\mathcal{I}$ captured at known camera poses $\mathcal{P}$ is used to learn a radiance field representation $\mathcal{G}$ of the scene.
$\mathcal{G}$ contains 2D Gaussian primitives  parameterized by center $\boldsymbol\mu_i\in \mathbb{R}^2$, covariance $\mathrm{\Sigma}_i$, base opacity $\alpha_i\in [0,\,1]$ and view-dependent color $\mathbf{c}_i$ denoted by Fourier transform coefficients operating on a query direction $\text{CH}_i(\theta)$. ~$\mathrm{\Sigma}_i=R_iS_iS_i^TR_i^T$ is decomposed into a rotation matrix $R_i$ and scaling matrix $S_i$.

\subsubsection{Extracting 3D Plenoptic Derivatives} \label{sec:flatland_derivatives}

The plenoptic derivatives measure sensitivity of a ray's intensity to shifts in its origin or direction.
\gls{PlenoCI} builds on $\nabla\mathscr{L} = [\mathscr{L}_x,\, \mathscr{L}_y,\, \mathscr{L}_{\theta}]$, where $\mathscr{L}_{*}=\partial \mathscr{L}\, /\, \partial*$ is the partial derivative for scalar $*$, or the Jacobian for vector~$*$.
\Crefrange{fig:flatland_dy}{fig:flatland_dt} show examples of $\mathscr{L}_*$ taken at a fixed $\theta$ slice.
We show that $\nabla\mathscr{L}$ is closed-form and analytic when $\mathscr{L}$ is regressed using \gls{3DGS}.

Radiance field representations can be queried to determine the color of a ray using volumetric rendering~\cite{mildenhallNeRFRepresentingScenes2020},
\begin{equation} \label{eqn:lr}
    \mathscr{L}(\mathbf{r})=\sum_i c_i\overline{\alpha}_i \prod_j^{i-1} (1-\overline{\alpha}_j),
\end{equation}
where $\overline{\alpha}$ is the effective opacity.

Prior \gls{3DGS} works show that closed-form analytic derivatives with respect to their parameters~\cite{kerbl3DGaussianSplatting2023} and camera pose~\cite{matsuki2024gaussian} can be computed.
Similarly, we can write derivatives with respect to $\mathbf{r}$.
Differentiating~\eqref{eqn:lr}, then using $\overline{\alpha}_i = \alpha_i \exp(\beta_i)$ and transmittance $T_i=\prod_j^{i-1}(1-\overline{\alpha}_j)$,
\begin{equation} \label{eqn:dlr_simplified}
    L_{*} = \sum_i \overline{\alpha}_iT_i
    \biggl( (c_i)_{*} + c_i\Bigl( (\beta_i)_{*} - \sum_j^{i-1}\frac{\overline{\alpha}_j (\beta_j)_{*}}{1-\overline{\alpha}_j} \Bigr) \biggr).
\end{equation}
Intuitively,~\eqref{eqn:dlr_simplified} contains a view-dependent color derivative term~$(c_i)_{*}$ describing the change in $\text{CH}_i(\theta)$ with $\mathbf{r}$, and a geometry term describing the change in fall-off $\beta_i$.
The former is trivial to compute, noting that $(c_i)_{x} = (c_i)_{y} = 0$.

For the latter, we follow~\cite{moenne20243d,wu20253dgut} by ray-tracing, \ie, for each ray, we evaluate intersecting Gaussians at the point of maximum response $\mathbf{p} = \mathbf{x} + \tau_{\text{max}}\mathbf{d}$, where $\mathbf{x}$ is the ray's origin and $\mathbf{d} = [\cos\theta,\,\sin\theta]^T$ is the direction.
Dropping the Gaussian subscripts for brevity, it can be shown that the Gaussian fall-off based on Mahalanobis distance $d_{\perp}$,  
\begin{equation} \label{eqn:Ai}
    \beta = -\frac{1}{2}d_{\perp}^2 = -\frac{1}{2}(\mathbf{p}-\boldsymbol\mu)^T\Sigma^{-1}(\mathbf{p}-\boldsymbol\mu),
\end{equation}
is minimized at the point of closest approach:
\begin{equation} \label{eqn:tau_max}
    \tau_{\text{max}} = \frac{(\boldsymbol\mu-\mathbf{x})^T\mathrm{\Sigma}^{-1}\mathbf{d}}{\mathbf{d}^T\mathrm{\Sigma}^{-1}\mathbf{d}} =
    \frac{-\mathbf{x}_g^T\mathbf{d}_g}{\mathbf{d}_g^T\mathbf{d}_g}.
\end{equation}

$\mathbf{x}_g=S^{-1}R^T(\mathbf{x}-\boldsymbol\mu)$ and $\mathbf{d}_g = S^{-1}R^T\mathbf{d}$ transforms $\mathbf{r}$ into the Gaussian's canonical frame.
We can apply the chain rule to derive analytic, closed-form expressions for $\beta_{*}$:
\begin{align}
    \beta_{\mathbf{x}} &= (\beta)_{d_{\perp}}\, (d_{\perp})_{\mathbf{x}_g}\, (\mathbf{x}_g)_{\mathbf{x}}, \label{eqn:flatland_ax} \\
    \beta_{\mathbf{\theta}} &= (\beta)_{d_{\perp}}\, (d_{\perp})_{\mathbf{d}_{g,\perp}}\, (\mathbf{d}_{g,\perp})_{\mathbf{d}_g}\, (\mathbf{d}_g)_{\mathbf{d}}\, (\mathbf{d})_{\theta}. \label{eqn:flatland_at}
\end{align}
$d_{\perp}$ is the orthogonal projection of $\mathbf{r}_g$ onto $\boldsymbol\mu$ in the canonical frame and $\mathbf{d}_{g,\perp}$ is the unit vector orthogonal to $\mathbf{d}_g$.

\subsubsection{The Plenoptic Structure Tensor} \label{sec:flatland_struct_tensor}

Edges and corners provide information about scene geometry and textures robust to illumination changes~\cite{shiGoodFeaturesTrack1994,laptev2005space}, making them desirable for image matching~\cite{image_matching_2005,ma2021image} and landmark detection~\cite{gil2010comparative,jeong_2006}.
Harris~\cite{harris1988combined} and Shi-Tomasi ~\cite{shiGoodFeaturesTrack1994} detect corners as points in an image $I$ with large intensity changes in two orthogonal directions.
We treat the plenoptic field in the same way, asking how a ray's color changes in $(x,y,\theta)$ to uncover occlusion edges and specularity.

The structure tensor smooths gradients over a window $w$,
\begin{equation} \label{eqn:struct_tensor_2d}
    M = \sum_{(x,y)} w(x,y)\begin{bmatrix}
        I_x^2 & I_x I_y \\
        I_y I_x & I_y^2
    \end{bmatrix},
\end{equation}
with corners detected where $M$ has two large eigenvalues.
We reformulate \eqref{eqn:struct_tensor_2d} with the angular dimension,
\begin{equation} \label{eqn:struct_tensor_3d}
    M=\sum_{(x,y,\theta)} w_{s}(x,y)\, w_{d}(\theta) \bigl(\nabla \mathscr{L}^*\bigr)\bigl(\nabla \mathscr{L}^{*}\bigr)^T.
\end{equation}
To control sensitivity to scene details, we smooth the analytic $\nabla \mathscr{L}$ using Gaussian windows $\omega_{s}$ and $\omega_{d}$ to construct $\nabla \mathscr{L}^*$, with $2\pi$ periodicity for $\omega_{d}$.
We use similar Gaussian windows $w_{s}$ and $w_{d}$ to aggregate outer products.
$M$ has 3 eigenvalues $\lambda_0\ge \lambda_1\ge\lambda_2\ge 0$, giving us 4 cases:
\begin{itemize}
    \item No large $\lambda$: a flat region.
    \item 1 large $\lambda$: a Lambertian texture or geometric edge where $\mathscr{L}$ is sensitive to shifts along one $(x,y,\theta)$ direction.
    \item 2 large $\lambda$: visually complex behaviors such as occlusion edges (\cref{fig:flatland_occlusion_crossings}) and specularity where $\mathscr{L}$ is sensitive to shifts along two orthogonal $(x,y,\theta)$ directions.
    \item 3 large $\lambda$: $\mathscr{L}$ is sensitive to any shift. This is highly unlikely in free space where rays do not change color along the direction of propagation.
\end{itemize}
For RGB, we write $\nabla\mathscr{L}^* = \begin{bmatrix} \nabla\mathscr{L}^*_R & \nabla\mathscr{L}^*_G & \nabla\mathscr{L}^*_B\end{bmatrix}$ to combine gradients across channels~\cite{dizenzo1986note} and apply~\eqref{eqn:struct_tensor_3d}. 

\subsubsection{PlenoCI in Flatland} \label{sec:flatland_plenoci}

We propose \gls{PlenoCI} as rays $\mathcal{R}$ where $M$ has two large eigenvalues, \ie, where $\lambda_1(\mathbf{r})\ge \tau_1$.
\Cref{fig:flatland_occlusion_crossings} shows regions with \gls{PlenoCI} \ie, rays observing an occlusion edge.

\begin{figure*}[tb]
    \centering
    \includegraphics[width=\linewidth]{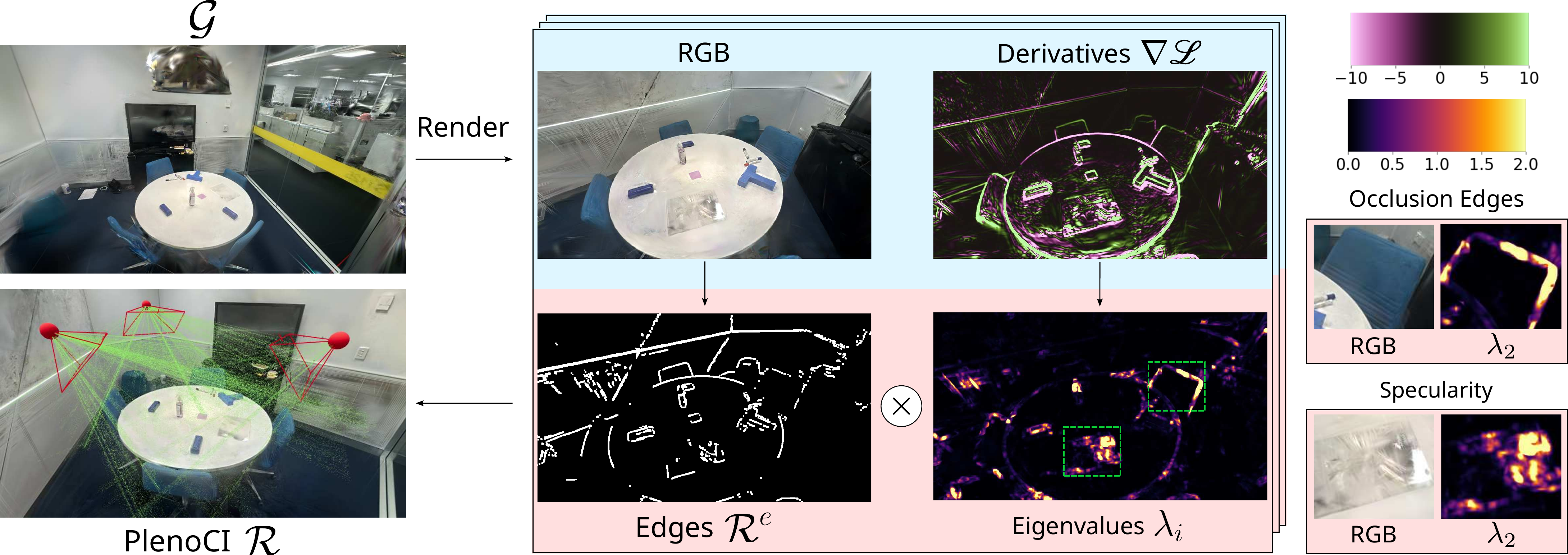}
    \caption{Our \gls{PlenoCI} feature extraction pipeline. Given a \gls{3DGS} representation $\mathcal{G}$, we sample analytic plenoptic derivatives $\nabla\mathcal{L}$ over a local light field to compute the structure tensor. The third eigenvalue $\lambda_2$ provides an informative signal for complex visual behaviors where existing features are unreliable, while ignoring Lambertian textures. We mask pixelwise $\lambda_2$ with an edge image (dilated for clarity) to yield \gls{PlenoCI} $\mathcal{R}^k$ from view $k$. Repeating this over selected views yields $\mathcal{R}.$}
    \label{fig:plenoci_overview}
\end{figure*}

\subsection{Generalizing to 5D} \label{sec:plenoci_generalizing}

Real scenes are 3D, calling for a 5D plenoptic field: $(x, y, z)$ for position and $(\phi,\theta)$ for direction.
\Cref{fig:plenoci_overview} shows our pipeline to extract 5D \gls{PlenoCI} from a \gls{3DGS} representation.
We explain key deviations from \Cref{sec:flatland} here and provide full details in~\Cref{sec:supp_derivations}.

In \eqref{eqn:dlr_simplified}, we highlighted two terms: the view-dependent color derivative $(c_i)_{\mathbf{r}}$ and geometry term $(\beta_i)_{\mathbf{r}}$.
Following~\cite{kerbl3DGaussianSplatting2023}, we model $c_i$ with \gls{SH}, queried using the unit vector $\hat{\mathbf{v}}$ from the camera to Gaussian's center, \ie, $c_i=\text{SH}_i(\hat{\mathbf{v}})$.
This leads to the surprising result that ray direction does \emph{not} affect $c_i$.
$(c_i)_{\mathbf{x}}$ can be computed with the chain rule.
For the geometry term, the Mahalanobis distance can be written as:
\begin{equation} \label{eqn:generalising_Ai_simplified}
    d_{\perp}^2 = \lVert \hat{\mathbf{d}}_g \times \mathbf{x}_g \rVert^2,
\end{equation}
where $\hat{\mathbf{d}}_g$ is the unit transformed direction.
Applying the chain rule in \eqref{eqn:flatland_ax}--\eqref{eqn:flatland_at} yields closed-form derivatives.

Following \eqref{eqn:struct_tensor_3d}, $M$ extends to a 5$\times$5 matrix.
To avoid sampling 5D volumes, we sample a local light field of analytic $\nabla\mathscr{L}$: a single pinhole camera provides $(\phi,\theta)$ samples while a 3$\times$3 camera array with fixed spacing also provides $(x,y,z)$ samples.
We use 2D Gaussian kernels for spatial and angular windows, and additionally scale spatial $\mathscr{L}_*$ by the median rendered depth before constructing $\nabla \mathscr{L}^*$.

The third eigenvalue $\lambda_2$ is sensitive to occlusion edges and specularity as \Cref{fig:plenoci_overview} shows.
Hence, it forms the basis of \gls{PlenoCI}: we threshold $\lambda_2 \ge \tau_2$, and mask this with a Canny edge map~\cite{canny_1986} to form $\mathcal{R}$.
We also find $\lambda_0$ is sensitive to edges, $\lambda_1$ to corners as \Cref{sec:supp_eig} shows.

%% file: 4_changes.tex
\section{Change Detection and Classification} \label{sec:change}

\begin{figure*}[t]
  \centering
  \includegraphics[width=\linewidth]{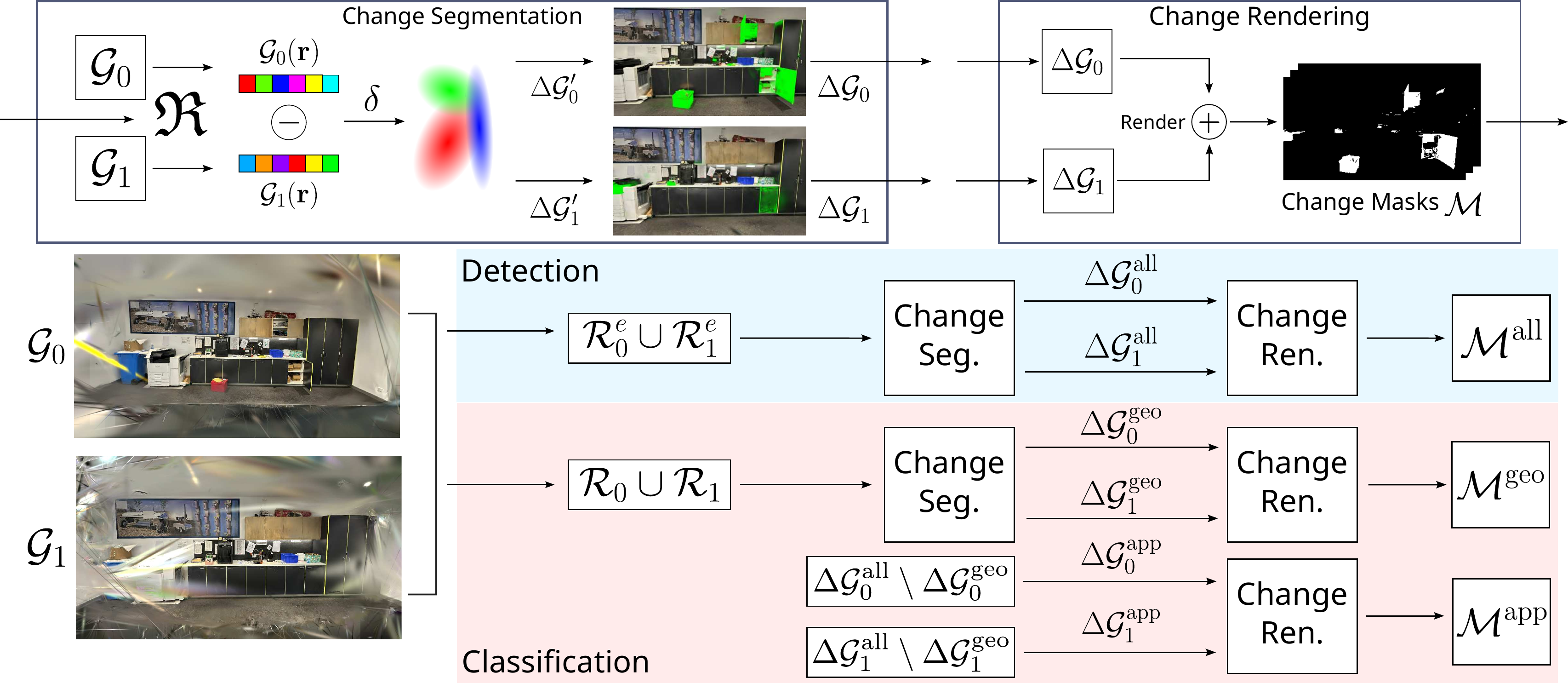}
  \caption{Our overall scene change detection and classification approach. Given instance-aware \gls{3DGS} scene representations $\mathcal{G}_0$ and $\mathcal{G}_1$, we aggregate per-ray change scores $\delta$ in the corresponding primitives. This yields a sparse set of anchors $\Delta\mathcal{G}'_t$ to propagate on changed instances. For change detection, we sample edge rays $\mathcal{R}^e$ to find changed Gaussians $\Delta\mathcal{G}^{\text{all}}_t$. For change classification, we leverage \gls{PlenoCI} $\mathcal{R}$ to find geometrically changed Gaussians $\Delta\mathcal{G}^{\text{geo}}_t$. These explicit representations can be used to render novel change masks with disambiguation between geometric and appearance changes.}
  \label{fig:generalising_overall}
\end{figure*}

\Cref{fig:generalising_overall} shows our overall approach.
Following prior work~\cite{Galappaththige_2025_CVPR}, we use \gls{SfM}~\cite{schonberger_sfm_2016} to register $\mathcal{I}_0$, $\mathcal{I}_1$ and find camera poses $\mathcal{P}_0$, $\mathcal{P}_1$ in a common coordinate frame.
We then construct instance-aware \gls{3DGS}~\cite{li2025instancegaussian} representations $\mathcal{G}_0$, $\mathcal{G}_1$.
Our pipeline aggregates ray differences at the instance level to detect and classify changes.
Given a set of rays, we accumulate change evidence in the corresponding primitives and threshold to yield sparse anchors $\Delta\mathcal{G}_t'$.
We propagate this information to related primitives using baked instance cues, yielding change representations $\Delta\mathcal{G}_t$ useful for rendering dense change masks.
For change detection, we use rays $\mathcal{R}^e$ observing image edges.
For change classification, we leverage \gls{PlenoCI} $\mathcal{R}$ to build geometric change representations $\Delta\mathcal{G}^{\text{geo}}_t$ from which geometric change masks are rendered.
Appearance change representations $\Delta\mathcal{G}^{\text{app}}_t=\Delta\mathcal{G}^{\text{all}}_t \setminus \Delta\mathcal{G}^{\text{geo}}_t$ can be used to render appearance change masks.

\subsection{Finding Changes} \label{sec:generalising_changes}

We seek representations $\Delta\mathcal{G}^{\text{all}}_t$ denoting changed primitives. 
For pose $k$, we select rays $\mathcal{R}^{e,k}_{t}$ on Canny edge~\cite{canny_1986} segments in the RGB render $I^k_{t}$.
To obtain reliable RGB and DINOv2 features~\cite{oquab2024dinov2}, we displace these boundary pixels towards the foreground using rendered depth gradients.
We follow~\cite{Galappaththige_2025_CVPR,galappaththige2025changes} to form per-ray change,
\begin{equation} \label{eqn:generalising_ray_change}
    \delta = w_{\text{rgb}}\delta_{\text{rgb}} + w_{\text{feat}}\delta_{\text{feat}},
\end{equation}
for all edge rays $\mathcal{R}^{e,k}_0\cup\mathcal{R}^{e,k}_1$.
$\delta_{\text{rgb}}$ is the L1 error and $\delta_{\text{feat}}$ is the cosine dissimilarity between feature embeddings.
We repeat this across all poses $k$ with fixed $w_{\text{rgb}}=w_{\text{feat}}=0.5$. 

To map rays to primitives, we consider the most dominant primitive along the ray, \ie, the primitive $i_t^*$ at timestep $t$ satisfying $i_t^*=\argmax_i \overline{\alpha}_i T_i$.
Mean-pooling $\delta$ in the frontmost of primitives $i_0^*$ and $i_1^*$ addresses two ambiguities:
1) primitives on the side of an edge of the changed object receive high $\delta$, and
2) whether the change is attributable to the frontmost primitive or one it occludes.
The latter resolves disocclusion cases (1) alone cannot distinguish.
This yields sparse anchors $\Delta\mathcal{G}_t'$ with change information.

To segment dense representations $\Delta\mathcal{G}_t$, we leverage learned per-Gaussian instance IDs.
Let $\mathcal{G}_t^{m}$ denote the set of primitives from time $t$ with instance ID $m$, and $\Delta\mathcal{G}_t^{'m}$ the set of anchor primitives on this instance.
$\mathcal{G}_t^{m}$ is marked as changed if the mean $\delta$ over $\Delta\mathcal{G}_t^{'m}$ exceeds $\tau_{\text{inst}}$.
We also impose an occupancy gate, where changed instances require an anchor fraction $\psi=\lvert \Delta\mathcal{G}_t^{'m}\rvert\,/\,\lvert\Delta\mathcal{G}^{m}\rvert \ge \tau_{\text{anc}}$.

Following~\cite{Galappaththige_2025_CVPR,galappaththige2025changes}, we render change views $M^{\text{all}}_{k,t}$ representing the change associated with time $t$ from novel view $k$. Combining the two timesteps $M^{\text{all}}_k = \text{max}(M^{\text{all}}_{k,0}, M^{\text{all}}_{k,1}) \ge \tau_{\text{mask}}=0.5$ yields the final mask.

\subsection{Classifying Changes with PlenoCI}

We use \gls{PlenoCI} as a proxy for geometry; occlusion edges shift when scene structure changes.
However, \gls{PlenoCI} also captures appearance properties such as specularity; we leave further disambiguation as future work.
Given \gls{PlenoCI} from both timesteps $\mathcal{R}_0\cup \mathcal{R}_1$, we compute change scores using~\eqref{eqn:generalising_ray_change}, and aggregate in the corresponding primitives,
\begin{equation}
    \delta(G_i)= \frac{\sum_{\mathbf{r}}w_{\text{geo}}(\mathbf{r})\,\delta(\mathbf{r})}{\sum_{\mathbf{r}}w_{\text{geo}}(\mathbf{r})}.
\end{equation}
$w_{\text{geo}}$ represents confidence of an occlusion edge change,
\begin{equation}
    w_{\text{geo}} = \lvert e_0-e_1\rvert,\quad e_t=\sigma\Bigl(\frac{\lambda_2-\tau_2}{T}\Bigr),
\end{equation}
where $T$ is a temperature hyperparameter, and $\sigma(\cdot)$ is the sigmoid activation function.

We similarly mark instances as a geometric change if the mean $\delta$ over its anchors exceeds $\tau^{\text{geo}}_{\text{inst}}$, and anchor fraction $\psi\ge\tau_{\text{anc}}$.
We also impose a minimum mass $\sum_{\mathbf{r}} w_{\text{geo}}(\mathbf{r})\ge \tau_{\text{geo}}=1$ to ensure sufficient geometric change support, and exclude instances not in $\Delta\mathcal{G}^{\text{all}}_t$.
This yields a set of geometric change primitives $\Delta\mathcal{G}^{\text{geo}}_t$.
All other changed primitives are marked as appearance-based, \ie, $\Delta\mathcal{G}^{\text{app}}_t=\Delta\mathcal{G}^{\text{all}}_t \setminus \Delta\mathcal{G}^{\text{geo}}_t$.
These representations can be used to render geometric and appearance change masks, respectively.

%% file: 5_results.tex
\section{Experiments} \label{sec:results}

We build instance-aware \gls{3DGS} representations with InstanceGaussian~\cite{li2025instancegaussian} while maintaining explicit ray-traced primitives~\cite{wu20253dgut} and densification~\cite{kheradmand20243d}.
We use gsplat~\cite{ye2025gsplat} with CUDA-accelerated kernels for real-time plenoptic derivatives.
All experiments were conducted on a single RTX PRO 6000 Blackwell Max-Q Workstation Edition GPU.
On average, training each scene takes 45 minutes while our \gls{SCD} and \gls{CC} pipeline takes 5 minutes.

We evaluate our method on CL-Splats~\cite{ackermann2025clsplats} and PASLCD~\cite{Galappaththige_2025_CVPR}, datasets featuring real-world scenes captured in-the-wild.
CL-Splats contains five scenes with simple geometric changes.
PASLCD has ten scenes with concurrent geometric and appearance changes, alongside two lighting variants each.
Ground truth labels for CL-Splats were annotated using the SceneDiff annotation tool~\cite{wu2026scenediff} and human-verified, while we use provided labels for PASLCD.

In our evaluations, we fix \gls{PlenoCI} extraction hyperparameters including smoothing windows, eigenvalue threshold $\tau_2$ and temperature $T$.
All pre-change images are used to train $\mathcal{G}_0$.
For $\mathcal{G}_1$, we hold out every 5${^\text{th}}$ image from training and evaluate \gls{SCD} on these unseen frames.
To build \gls{PlenoCI}, we sample 20 post-change training views---this corresponds to all training views for PASLCD, and a subset for CL-Splats selected using farthest point sampling~\cite{eldar1997fps}.
We use a grid search to select per-scene instance thresholds $\tau_{\text{inst}}$, $\tau_{\text{inst}}^{\text{geo}}$, and minimum anchor fraction $\tau_{\text{anc}}$.

\subsection{Binary Change Detection Results} \label{sec:results_binary}

We compare our approach to state-of-the-art methods: CYWS~\cite{sachdeva2023change}, GeSCF~\cite{Kim_2025_CVPR}, SceneDiff~\cite{wu2026scenediff}, 3DGS-CD~\cite{lu3DGSCD3DGaussian2025}, MV3DCD~\cite{Galappaththige_2025_CVPR}, OSCD~\cite{galappaththige2025changes} and GS-Diff~\cite{galappaththige2026gsdiff}.
For fair comparison, we benchmark OSCD~\cite{galappaththige2025changes} in the offline setting and evaluate pairwise methods~\cite{sachdevaChangeYouWant2023,Kim_2025_CVPR} by rendering aligned views using vanilla \gls{3DGS}~\cite{kerbl3DGaussianSplatting2023}.
To assess change masks, we follow prior \gls{SCD} work~\cite{linRobustSceneChange2024,lu3DGSCD3DGaussian2025,park2021changesim,Galappaththige_2025_CVPR} and report \gls{mIoU} and F1 score for the changed class.

\begin{figure*}[tb]
  \centering
  \includegraphics[width=\linewidth]{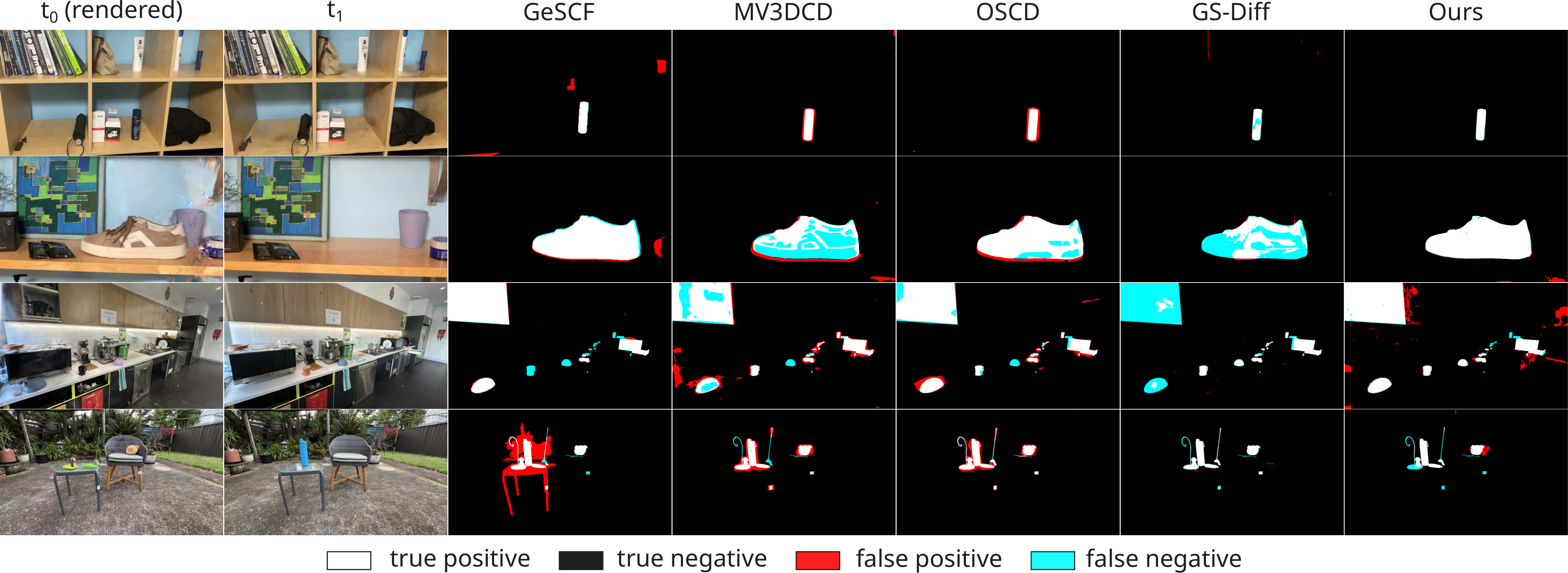}
  \caption{Qualitative comparison of binary change masks for CL-Splats~\cite{ackermann2025clsplats} (rows 1--2) and PASLCD~\cite{Galappaththige_2025_CVPR} (rows 3--4). Pairwise approaches such as GeSCF~\cite{Kim_2025_CVPR} are susceptible to spurious detections. Our instance-aware \gls{3DGS} approach outperforms all baselines on CL-Splats, and is competitive to state-of-the-art on the challenging PASLCD benchmark.}
  \label{fig:results_binary}
\end{figure*}

\Cref{tab:results_cd} summarizes binary \gls{SCD} results.
Our approach surpasses all baselines on CL-Splats, making a 25.7\% \gls{mIoU} improvement over the strongest competitor, while we closely match the strongest performance on PASLCD.
This stems from leveraging instance-aware 3D representations.
As \Cref{fig:results_binary} shows, our approach produces complete segmentations for singular changed objects in CL-Splats, while methods lacking this information~\cite{wu2026scenediff,Galappaththige_2025_CVPR,galappaththige2025changes} provide incomplete masks.
PASLCD contains more cluttered scenes where instance segmentation quality can limit our performance; we perform oracle experiments in \Cref{sec:results_ablation}. 

\begin{table}[t]
    \centering
    \caption{Binary \gls{SCD} results averaged over CL-Splats~\cite{ackermann2025clsplats} and PASLCD~\cite{Galappaththige_2025_CVPR}. PASLCD baselines are sourced from~\cite{galappaththige2025changes,wu2026scenediff}. Our method achieves state-of-the-art performance on CL-Splats, and is competitive with the strongest PASLCD baselines. The \first{first}, \second{second}, and \third{third} best performances are highlighted.}
    \label{tab:results_cd}
        \begin{tabular}{lcccc}\toprule
            \multirow{2}{*}{Method} & \multicolumn{2}{c}{CL-Splats~\cite{ackermann2025clsplats}} & \multicolumn{2}{c}{PASLCD~\cite{Galappaththige_2025_CVPR}} \\
            \cmidrule(lr){2-3} \cmidrule(lr){4-5}
            & mIoU $\uparrow$ & F1 $\uparrow$ & mIoU $\uparrow$ & F1 $\uparrow$ \\ \midrule
            CYWS~\cite{sachdeva2023change} & 0.495 & 0.640 & 0.273 & 0.398 \\
            GeSCF~\cite{Kim_2025_CVPR} & 0.692 & 0.793 & 0.477 & 0.611 \\
            SceneDiff~\cite{wu2026scenediff} & 0.334 & 0.433 & 0.473 & -- \\
            3DGS-CD~\cite{lu3DGSCD3DGaussian2025} & 0.700 & 0.792 & 0.209 & 0.339 \\
            MV3DCD~\cite{Galappaththige_2025_CVPR} & 0.634 & 0.752 & 0.478 & 0.628 \\
            OSCD~\cite{galappaththige2025changes} & \second{0.756} & \second{0.845} & \second{0.552} & \second{0.694} \\
            GS-Diff~\cite{galappaththige2026gsdiff} & \third{0.724} & \third{0.829} & \first{0.630} & \first{0.745} \\
            Ours & \first{0.951} & \first{0.975} & \third{0.538} & \third{0.688} \\ \bottomrule
        \end{tabular}
\end{table}

\subsection{Ablations} \label{sec:results_ablation}

\begin{table}[tb]
    \centering
    \caption{Change scoring ablation averaged over PASLCD~\cite{Galappaththige_2025_CVPR}. Incorporating all elements provides the best performance.}
    \label{tab:ablation_change}
    \begin{tabular}{lcc} \toprule
        Method & \gls{mIoU} $\uparrow$ & F1 $\uparrow$ \\ \midrule
        No RGB          & 0.530 & 0.671 \\
        No Semantics    & 0.469 & 0.614 \\
        No Occupancy   & 0.514 & 0.658 \\
        \textbf{Ours}  & \textbf{0.538} & \textbf{0.688} \\ \bottomrule
    \end{tabular}
\end{table}
In \Cref{tab:ablation_change}, we ablate our change scoring using identical reconstructions and rays $\mathcal{R}^e$.
As~\cite{Galappaththige_2025_CVPR,galappaththige2025changes} find, RGB and semantics play complementary roles.
Occupancy gating inductively biases our \gls{SCD} pipeline towards localized changes, preventing a small number of false positive anchors from propagating through large instances.
Large changed instances with sufficient evidence are preserved.

Since instance segmentation is upstream and orthogonal to \gls{PlenoCI}, our \gls{SCD} pipeline directly benefits from improvements to the segmenter.
PASLCD contains small changes which may be localized by the anchors but undersegmented in the \gls{3DGS} representation, with qualitative examples in \Cref{sec:supp_seg}.
We validate this by substituting learned instance IDs~\cite{li2025instancegaussian} with ground truth oracles.
Concretely, we run connected components analysis with 8-connectivity on ground truth masks.
We lift blob pixels to 3D using the most dominant primitive (\cref{sec:generalising_changes}), and group primitives by spatial proximity.
The lifted groups are merged by intersection over union to form the oracles.
Since unchanged objects do not receive an oracle, change \emph{recall} is the main metric.
As \Cref{tab:ablation_oracle} shows, making this substitution with fixed anchors increases recall by 20\% while leaving precision stable, confirming our hypothesis.

\begin{table}[tb]
    \centering
    \caption{\gls{SCD} results averaged over PASLCD~\cite{Galappaththige_2025_CVPR} varying instance segmentations. Oracle derives instances from ground truth masks while we use learned instances~\cite{li2025instancegaussian}. Recall is the main metric, improving by 20\% with oracle instances as the learned instances undersegment small changed objects.}
    \label{tab:ablation_oracle}
    \begin{tabular}{lcccc}\toprule
        Method & \gls{mIoU} $\uparrow$ & F1 $\uparrow$ & Precision $\uparrow$ & Recall $\uparrow$ \\ \midrule
        Ours & 0.538 & 0.688 & \textbf{0.712} & 0.691 \\
        Oracle & \textbf{0.596} & \textbf{0.731} & 0.667 & \textbf{0.839} \\ \bottomrule
    \end{tabular}
\end{table}

We also compare querying only edge rays with all rays from the sample views in \Cref{sec:supp_ray_selection}.
Querying all rays yields comparable SCD performance, dropping \gls{mIoU} by 0.021 while considerably increasing compute costs.

\subsection{Multi-class Change Classification Results} \label{sec:results_multi}

\begin{figure*}[tb]
  \centering
  \includegraphics[width=\linewidth]{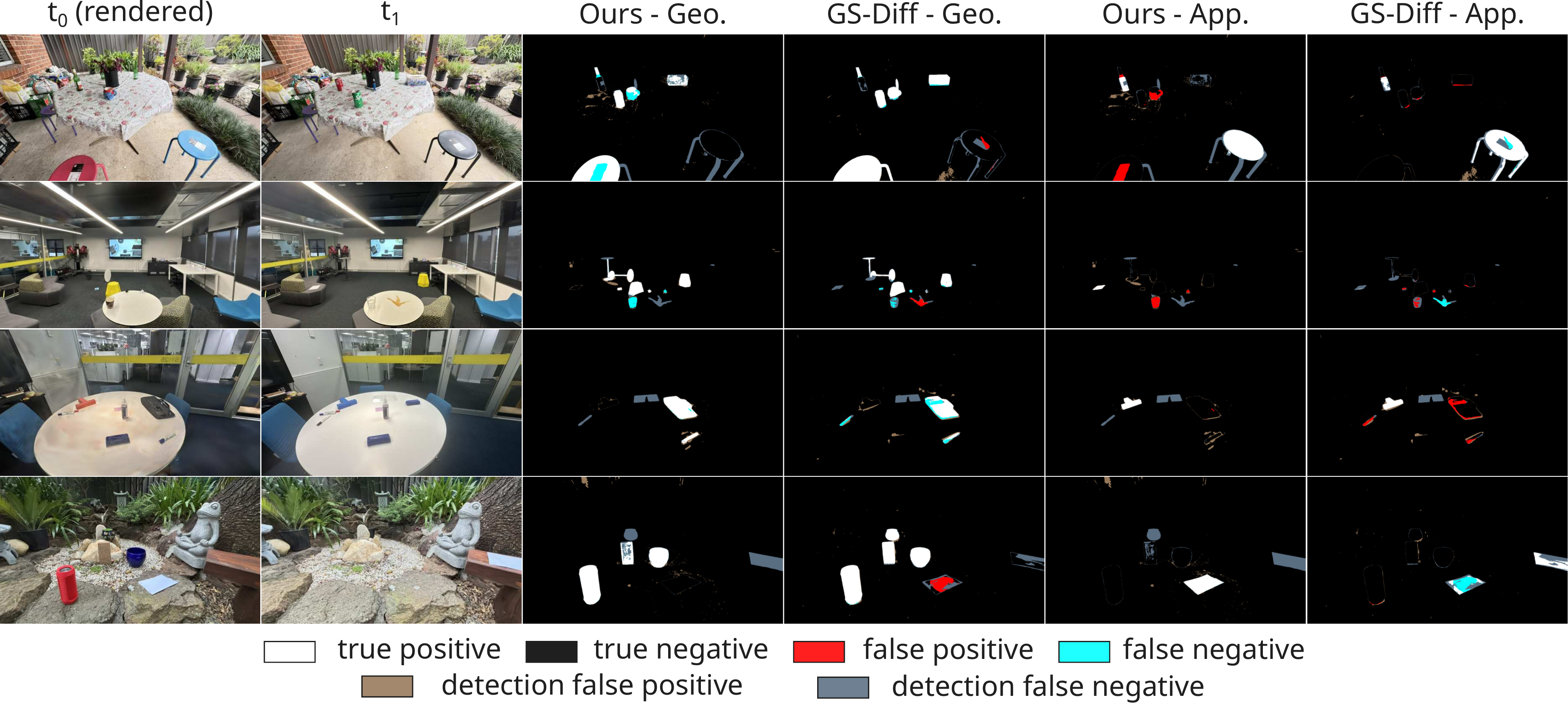}
  \caption{Qualitative change classification results for PASLCD~\cite{Galappaththige_2025_CVPR}. Top to bottom: Porch, Lounge, Meeting Room, Zen. Our \gls{PlenoCI} feature enables building explicit geometry and appearance-based change representations for rendering multi-class change masks.}
  \label{fig:results_multi}
\end{figure*}

\begin{table}[tb]
    \centering
    \caption{\gls{CC} performance averaged over PASLCD~\cite{Galappaththige_2025_CVPR}. The best performance excluding Oracle is \textbf{bolded}. Oracle uses instances derived from ground truth masks (\cref{sec:results_ablation}), driving improvements in recall of both classes. Ground truth class imbalance favoring Geo amplifies App classification errors.}
    \label{tab:results_multi}
    \begin{tabular}{lccc} \toprule
        Metric & GS-Diff~\cite{galappaththige2026gsdiff} & Ours & Oracle \\ \midrule
        Balanced Accuracy $\uparrow$ & \textbf{0.792} & 0.735 & 0.851 \\
        Geo Precision $\uparrow$ & 0.970 & \textbf{0.985} & 0.970 \\
        Geo Recall $\uparrow$ & \textbf{0.896} & 0.745 & 0.871 \\
        App Precision $\uparrow$ & \textbf{0.478} & 0.431 & 0.463 \\
        App Recall $\uparrow$ & 0.687 & \textbf{0.725} & 0.830 \\
        \bottomrule
    \end{tabular}
\end{table}

To physically ground changes, we classify them as geometric (Geo) or appearance-based (App).
We evaluate multi-class performance on PASLCD~\cite{Galappaththige_2025_CVPR} using human-annotated Geo or App classifications from~\cite{galappaththige2026gsdiff}.
We report balanced accuracy~\cite{brodersen2010balanced} and per-class precision and recall.

\Cref{tab:results_multi} and \Cref{fig:results_multi} present our \gls{CC} results, using \gls{PlenoCI} as a proxy for geometry.
Our approach is competitive with GS-Diff.
App precision is low across methods due to class imbalance; 84\% of change pixels are Geo, amplifying App errors~\cite{galappaththige2026gsdiff}.
\gls{PlenoCI} conflates occlusion edges with specularity and reflections.
Conversely, object replacements (\eg, the mug and pen in Porch) can create overlapping \gls{PlenoCI}, misclassifying these Geo changes as App.
Upstream segmentation also affects \gls{CC}; using oracle instances (\cref{sec:results_ablation}) improves recall for both classes.

\subsection{Robustness to Reconstruction Ambiguity} \label{sec:results_ambiguity}

We take an orthogonal approach to concurrent work.
GS-Diff~\cite{galappaththige2026gsdiff} directly compares primitives, mitigating reconstruction ambiguity with explicit drift modeling while assuming geometrically accurate representations.
We derive \gls{PlenoCI} from the plenoptic field, remaining robust to underconstrained primitives by construction.
To evaluate this, we independently optimize two \gls{3DGS} models of a static scene using all pre-change images, reporting all detections as false positives.
For our method, we use the same $(\tau_{\text{inst}},\tau_{\text{anc}},\tau_{\text{inst}}^{\text{geo}})$ from Sections~\ref{sec:results_binary} and~\ref{sec:results_multi}.
As \Cref{tab:results_ambiguity} shows, our false positive rates are consistently lower.
Our approach is also robust to reconstruction ambiguity induced by varying training view overlaps and coverage levels (\cref{sec:supp_ambiguity}).

\begin{table}[tb]
    \centering
    \caption{False positive rate from comparing two independently optimized \gls{3DGS} models of an unchanged scene, averaged over PASLCD~\cite{Galappaththige_2025_CVPR}. Scenes with real changes contain 3.51\% change pixels on average, ranging from 0.17--20.12\%.}
    \label{tab:results_ambiguity}
    \begin{tabular}{lccc} \toprule
        Method & Binary (\%) $\downarrow$ & Geo (\%) $\downarrow$ & App (\%) $\downarrow$ \\ \midrule
        GS-Diff~\cite{galappaththige2026gsdiff} & 0.389 & 0.064 & 0.325 \\
        Ours & \textbf{0.004} & \textbf{0.001} & \textbf{0.003} \\ \bottomrule
    \end{tabular}
\end{table}

%% file: 6_conclusion.tex
\section{Conclusion} \label{sec:conclusion}

We introduced \gls{PlenoCI}, a novel feature that explicitly captures complex visual behaviors discarded by existing methods, while being robust to underconstrained representations.
We showed that using \gls{3DGS} to regress the plenoptic field yields closed-form analytic derivatives.
In this work, we demonstrated the utility of \gls{PlenoCI} for distinguishing between geometric and appearance-based changes.
Future work could also conduct instance segmentation in plenoptic space, propagating change evidence along related rays.
Beyond \gls{SCD}, we believe this work lays a foundation for robust autonomous perception in visually complex environments.
\gls{PlenoCI} could be applicable in tasks such as localization, navigation and map stitching, while the underlying plenoptic derivatives could be informative for visual odometry and uncertainty estimation.

%% file: 7_acknowledgements.tex
\section*{Acknowledgements}

This research was supported in part through the NVIDIA Academic Grant Program and by the ARC Research Hub in Intelligent Robotic Systems for Real-Time Asset Management (IH210100030).
J.L. acknowledges ongoing support by the Australian Government Research Training Program (RTP) Scholarship. C.J., N.S., and D.M. acknowledge ongoing support from the QUT Centre for Robotics.

%% file: supplementary.tex
\clearpage
\setcounter{page}{1}
\maketitlesupplementary
\setcounter{equation}{0}
\renewcommand{\theequation}{S\arabic{equation}}

\section{Plenoptic Gradients Derivation} \label{sec:supp_derivations}

\subsection{Volumetric Rendering}

The color of a ray $\mathbf{r}$ can be computed using volumetric rendering as per~\eqref{eqn:lr},
\begin{equation} \label{eqn:supp_lr}
    \mathscr{L}(\mathbf{r}) = \sum_i c_i \overline{\alpha}_i \prod_j^{i-1}(1-\overline{\alpha}_j).
\end{equation}
Our goal is to differentiate this with respect to $\mathbf{r}$.
Let $C_i=c_i\overline{\alpha}_i$, $T_i=\prod_j^{i-1}(1-\overline{\alpha}_j)$ and $\mathscr{L}_i=C_iT_i$.
Differentiating with the product rule yields:
\begin{equation} \label{eqn:supp_li}
    (\mathscr{L}_i)_* = (C_i)_*T_i + C_i(T_i)_*.
\end{equation}
Taking the natural logarithm of $T_i$, differentiating and rearranging gives:
\begin{equation} \label{eqn:supp_dTi}
    (T_i)_* = -T_i\sum_j^{i-1}\frac{(\overline{\alpha}_j)_*}{1-\overline{\alpha}_j}.
\end{equation}
Substituting~\eqref{eqn:supp_dTi} and $(C_i)_*=(c_i)_*\overline{\alpha}_i + c_i(\overline{\alpha}_i)_*$ into~\eqref{eqn:supp_lr} and~\eqref{eqn:supp_li} yields~\eqref{eqn:dlr_simplified}.

\subsection{Ray-Gaussian Intersection}

We now consider derivatives of the geometric fall-off term $(\beta_i)_*$.
Dropping the Gaussian subscripts for brevity, we can write from~\eqref{eqn:Ai}:
\begin{equation}
    \beta_* = -\frac{1}{2}(d_{\perp}^2)_*,
\end{equation}
where $d_{\perp}$ is the Mahalanobis distance given by~\eqref{eqn:generalising_Ai_simplified}:
\begin{equation} \label{eqn:supp_Ai_simplified}
    d_{\perp}^2 = \lVert \hat{\mathbf{d}}_g \times \mathbf{x}_g \rVert^2.
\end{equation}
$\mathbf{x}_g$ and $\hat{\mathbf{d}}_g$ represent the ray transformed into canonical Gaussian space.
Using $\lVert \hat{\mathbf{d}}_g \rVert =1$, we can write:
\begin{align}
   (d_{\perp}^2)_{\mathbf{x}_g} &=  2\bigl(\mathbf{x}_g - (\mathbf{x}_g \cdot \hat{\mathbf{d}}_g)\hat{\mathbf{d}}_g\bigr), \\
   (d_{\perp}^2)_{\hat{\mathbf{d}}_g} &=  2\bigl(\lVert \mathbf{x}_g \rVert^2 \hat{\mathbf{d}}_g - (\mathbf{x}_g \cdot \hat{\mathbf{d}}_g)\mathbf{x}_g\bigr).
\end{align}

Given a primitive with center $\boldsymbol{\mu}$, scale matrix $S$ and rotation matrix $R$, the world-to-canonical projection of a ray with origin $\mathbf{x}$ and direction $\mathbf{d}$ is given by:
\begin{align}
    \mathbf{x}_g &= S^{-1}R^T(\mathbf{x}-\boldsymbol\mu), \\
    \mathbf{d}_g &= S^{-1}R^T\mathbf{d},
\end{align}
The spatial derivatives can be evaluated as:
\begin{equation}
    \begin{bmatrix} \beta_x & \beta_y & \beta_z \end{bmatrix} = -\frac{1}{2} (d_{\perp}^2)_{\mathbf{x}_g}S^{-1}R^T I_3,
\end{equation}
where $I_3$ is the identity matrix.
The angular term requires the chain rule for normalization, world-to-canonical projection and the Jacobian $J_{\text{ang}}$.
We parameterize direction with spherical coordinates: let $\phi\in [-\pi,\,\pi]$ represent the ray's azimuth in the reference $xy$ plane, and $\theta\in [-\pi/2,\, \pi/2]$ be the elevation from this plane.
Then $J_{\text{ang}}$ is
\begin{equation}
    J_{\text{ang}}=\begin{bmatrix} \mathbf{d}_{\phi} \\ \mathbf{d}_{\theta} \end{bmatrix}^T  = \begin{bmatrix} -\sin\phi\cos\theta & -\cos\phi\sin\theta \\
    \cos\phi\cos\theta & -\sin\phi\sin\theta \\
    0 & \cos\theta \end{bmatrix}.
\end{equation}
Therefore, the angular derivatives can be evaluated as:
\begin{equation}
    \begin{bmatrix} \beta_{\phi} & \beta_{\theta} \end{bmatrix} = 
    -\frac{1}{2} (d_{\perp}^2)_{\hat{\mathbf{d}}_g} (\hat{\mathbf{d}}_g)_{\mathbf{d}_g}(\mathbf{d}_g)_{\mathbf{d}}\,J_{\text{ang}}.
\end{equation}

Each of these terms can be accumulated per-Gaussian and alpha-blended following standard ray-traced 3DGS rasterization~\cite{moenne20243d}.
To prevent numerical instability of the derivatives for flat primitives, we add $\varepsilon=10^{-3}$ to the scaling matrix during all forward and backward passes,
\begin{equation}
    S^{-1}_{i,i}= \frac{1}{s_i+\varepsilon}.
\end{equation}
We find this causes a small drop in render quality as it imposes a lower bound on a primitive's scale.

\section{Additional Visualizations} \label{sec:supp_eig}
\begin{figure*}[tb]
    \centering
    \includegraphics[width=\linewidth]{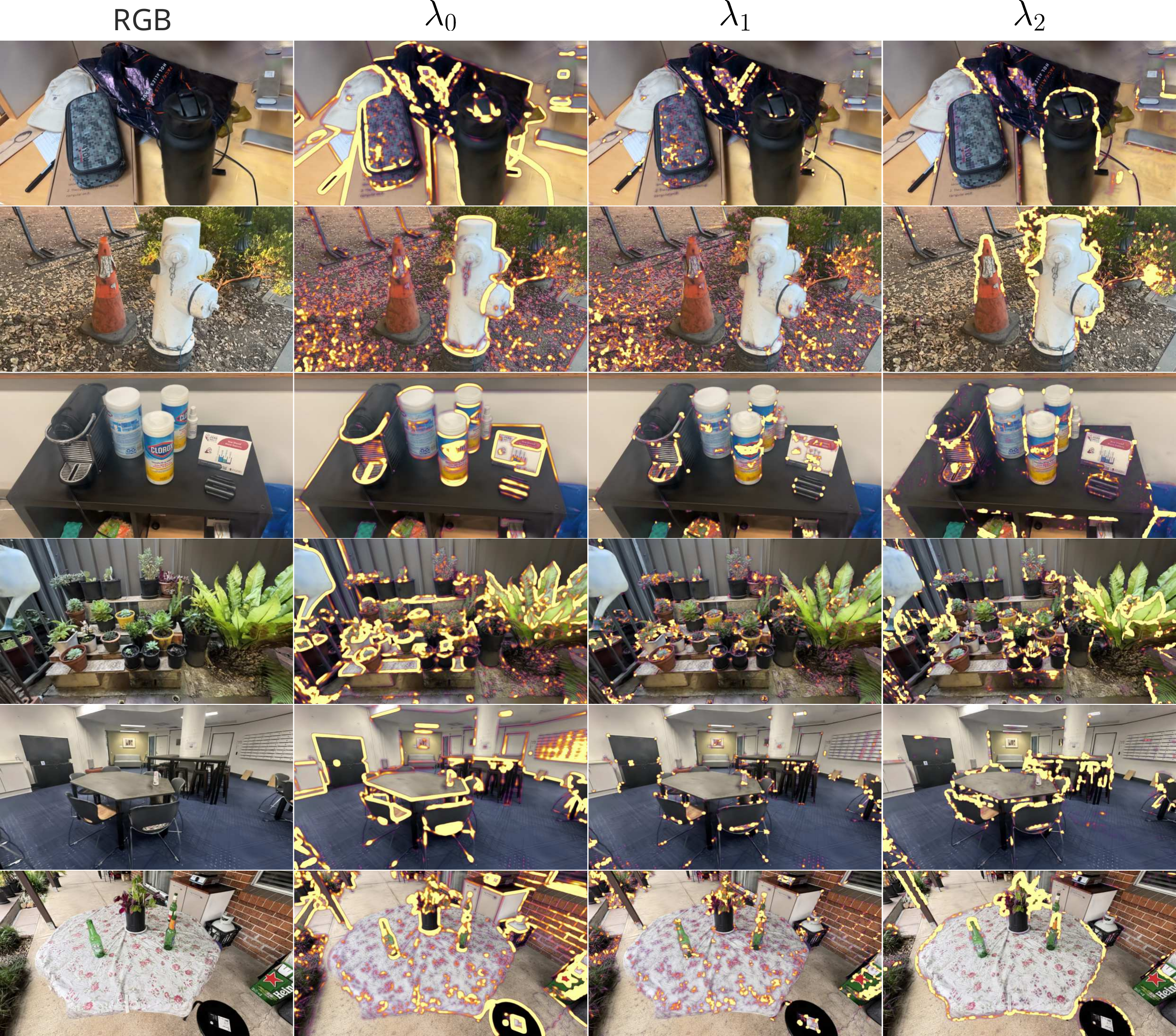}
    \caption{Pixelwise eigenvalue signals derived from the plenoptic structure tensor. Given an RGB view (column 1), we sample analytic plenoptic derivatives over a local light field to build pixelwise structure tensors $M$. The eigenvalues $\lambda_i$ of $M$ provide informative signals of scene structures, such as edges (column 2), corners (column 3) and occlusion edges as well as specular highlights (column 4).}
    \label{fig:supp_eig}
\end{figure*}

\Cref{fig:supp_eig} presents additional visualizations of per-pixel eigenvalues for the plenoptic structure tensor $M$ developed in \Cref{sec:plenoci_generalizing}.
In a 3D world, $M$ has 5 eigenvalues $\lambda_0\ge\lambda_1\ge\ldots\ge\lambda_4\ge 0$.
We find that $\lambda_0$ is sensitive to edges, $\lambda_1$ to corners, and $\lambda_2$ to occlusion edges, specularity and reflections.

\subsection{Instance Segmentation} \label{sec:supp_seg}

\begin{figure*}[tb]
    \centering
    \includegraphics[width=\linewidth]{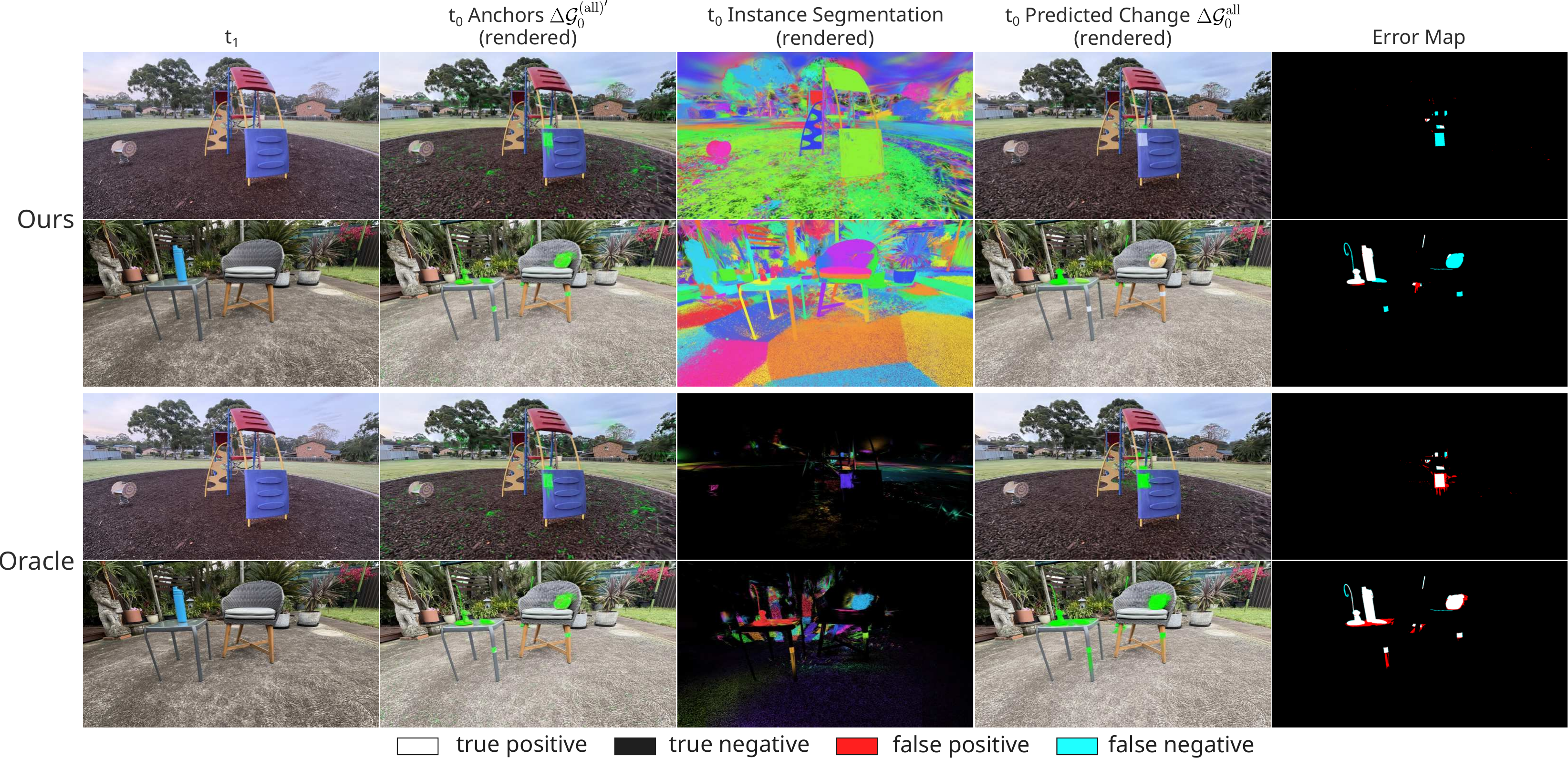}
    \caption{Propagation of change information from anchors $\Delta\mathcal{G}'_t$ (column 1, highlighted in green) to produce dense change representations $\Delta\mathcal{G}_t$ (column 3, highlighted in green) depends on the quality of the instance-aware \gls{3DGS} (column 2). While our anchors correctly identify changed objects, upstream undersegmentation (rows 1--2) can cause our instance gating to attenuate these anchors, creating false negative predictions (column 4). With oracle segmentations (rows 3--4), our predicted changes are considerably more accurate.}
    \label{fig:supp_seg}
\end{figure*}

Our approach uses InstanceGaussian~\cite{li2025instancegaussian} to build an instance-aware \gls{3DGS} representation.
These instance cues enable change information from sparse anchors $\Delta\mathcal{G}'_t$ to propagate to all other primitives on the changed object.
Thus, our \gls{SCD} and \gls{CC} performance depends on the instance segmentation quality.
As noted in \Cref{sec:results_binary}, PASLCD contains numerous small changed objects.
\Cref{fig:supp_seg} shows two examples of undersegmentation (rows 1--2) where the anchors (column 2) correctly localize changes, but are prevented from propagating by our instance gating due to an undersegmented representation (column 3), producing false negatives (columns 4--5).
Substituting with oracle instances (rows 3--4) mitigates this, greatly reducing false negative predictions.

\subsection{Ray Selection Comparison} \label{sec:supp_ray_selection}

\begin{table}[tb]
    \centering
    \caption{Comparison of ray selection approaches averaged over all scenes in PASLCD~\cite{Galappaththige_2025_CVPR} using 20 sample views each. Floating point operations reported for Extraction (Ext) do not include rasterization, which is shared across both methods. Retaining only edge rays (Ours) marginally improves \gls{SCD} performance, but substantially reduces compute costs for change scoring (Sco).}
    \label{tab:supp_ablation_ray_selection}
    \begin{tabular}{lcc} \toprule
        Metric & All Rays & Edge Rays (Ours) \\ \midrule
        \gls{mIoU} $\uparrow$ & 0.517 & \textbf{0.538} \\
        F1 $\uparrow$ & 0.661 & \textbf{0.688} \\ \midrule
        Ext TFLOPs $\downarrow$ & 35.0 & \textbf{34.9} \\
        Ext VRAM (GB) $\downarrow$ & 10.3 & \textbf{5.13} \\ \midrule
        Sco GFLOPs $\downarrow$ & 1.11 & \textbf{0.043} \\
        Sco VRAM (GB) $\downarrow$ & 3.64 & \textbf{1.92} \\ 
        \bottomrule
    \end{tabular}
\end{table}
We compare ray selection approaches in \Cref{tab:supp_ablation_ray_selection}, using the same \gls{3DGS} reconstructions for fair comparison.
Querying all rays in the sample views yields comparable \gls{SCD} performance.
During ray extraction, most \gls{FLOPs} occur during DINOv2 inference, requiring entire frames regardless of the ray selection approach.
Substantial compute savings occur during downstream change scoring: with edge rays occupying on average 3.6\% of the frame, a proportional drop in \gls{FLOPs} follows.

\subsection{Robustness to Reconstruction Ambiguity} \label{sec:supp_ambiguity}

We follow the protocol in~\Cref{sec:results_ambiguity}, reporting all changes detected between the \gls{3DGS} reconstructions as false positives.
We investigate two conditions: 1) varying the fraction of overlapping views in each model's training views, and 2) reducing the coverage of views provided to one model.
For our method, we use the same $(\tau_{\text{inst}},\tau_{\text{anc}}, \tau_{\text{inst}}^{\text{geo}})$ from the main results (\cref{sec:results_binary}, \cref{sec:results_multi}).

\begin{table}[htbp]
    \centering
    \caption{False positive rate from comparing two independently optimized \gls{3DGS} reconstructions of the Cantina scene from PASLCD~\cite{Galappaththige_2025_CVPR} under no changes. Under varying training view overlaps and levels of coverage, our method consistently reports fewer false positives than GS-Diff~\cite{galappaththige2026gsdiff}.}
    \label{tab:supp_ambiguity}
    \begin{tabular}{lcc} \toprule
        Condition & Ours (\%) $\downarrow$ & GS-Diff (\%) $\downarrow$\\
        \midrule
        \multicolumn{3}{l}{\textit{View Overlap}} \\
        100\% (full vs. full) & 0.000 & 0.487 \\
        50\% (half vs. half) & 0.003 & 0.397 \\
        0\% (half vs. half)  & 0.020 & 0.446 \\
        \midrule
        \multicolumn{3}{l}{\textit{Coverage}} \\
        Full vs. full & 0.000 & 0.487 \\
        Full vs. half & 0.002 & 0.446 \\
        Full vs. quarter & 0.142 & 0.442 \\ \bottomrule
    \end{tabular}
\end{table}

\begin{figure*}[tb]
    \centering
    \includegraphics[width=\linewidth]{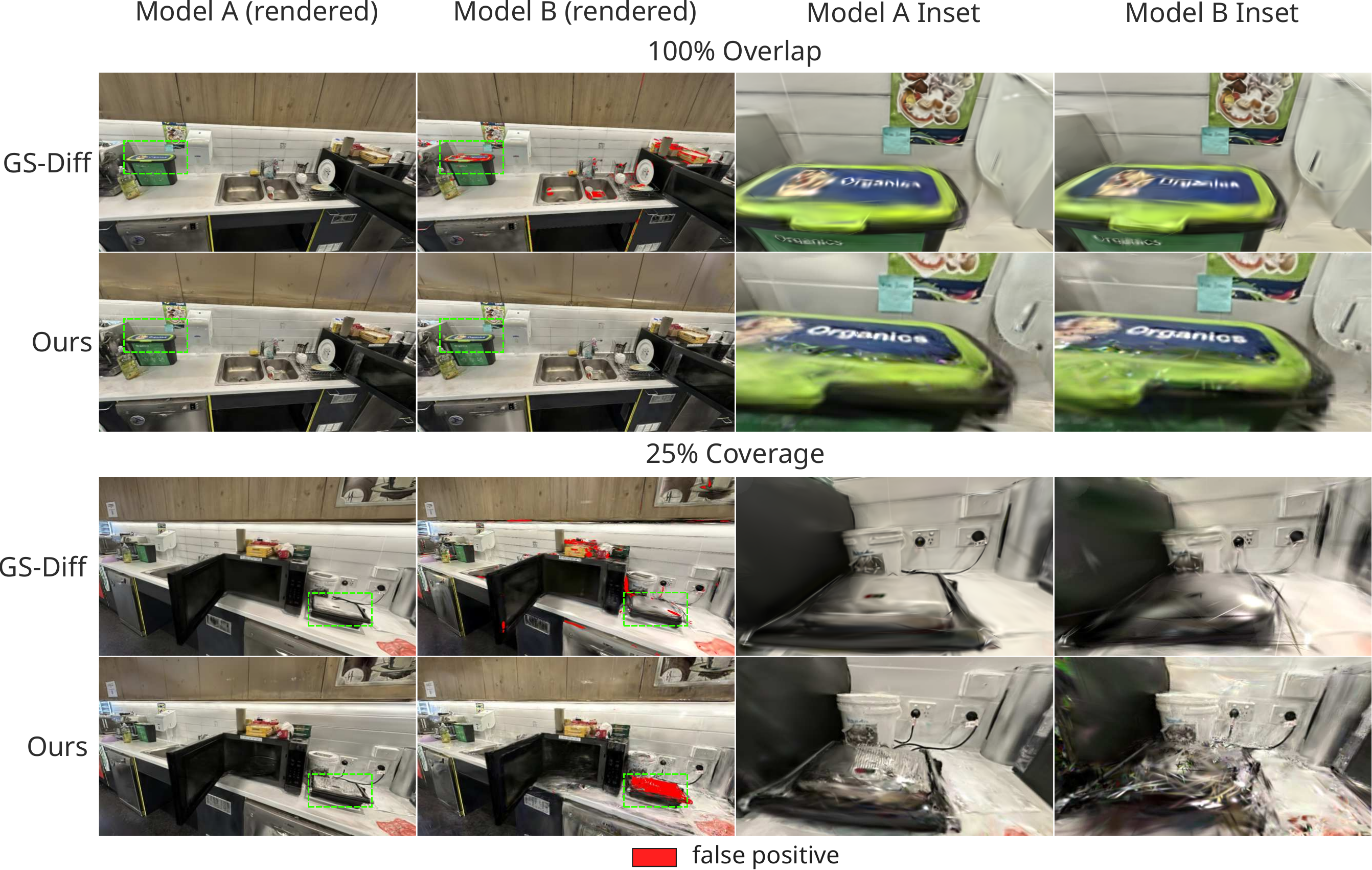}
    \caption{Comparison of independently optimized \gls{3DGS} models of the Cantina scene from PASLCD without changes. At 100\% view overlap, both models receive all training views. GS-Diff~\cite{galappaththige2026gsdiff} reports a false positive rate of 0.487\% as it directly compares primitives (inset), while we correctly report no changes. At 25\% view coverage, Model B receives 5 out of 20 training views. Poor reconstruction quality causes our approach to detect false positives, albeit at a lower rate than GS-Diff.}
    \label{fig:supp_ambiguity}
\end{figure*}

\Cref{tab:supp_ambiguity} summarizes our results for Cantina from PASLCD~\cite{Galappaththige_2025_CVPR}, a challenging scene with fine details and view-dependent objects prone to reconstruction ambiguity.
As \Cref{fig:supp_ambiguity} shows, our approach consistently reports fewer false positive changes than GS-Diff, stemming from our comparisons in plenoptic rather than primitive space.

Our false positive rate increases under severe coverage asymmetry (\cref{fig:supp_ambiguity} rows 3--4), but remains below GS-Diff.
With very few views to constrain the optimization, reconstruction quality drops as fine details are lost, causing spurious change cues from ray differencing to leak into our pipeline.
GS-Diff uses Fisher information to model observability as a way to absorb this reconstruction ambiguity, allowing a consistent false positive rate to be maintained.

%% file: main.bbl
\begin{thebibliography}{78}
\providecommand{\natexlab}[1]{#1}
\providecommand{\url}[1]{\texttt{#1}}
\expandafter\ifx\csname urlstyle\endcsname\relax
  \providecommand{\doi}[1]{doi: #1}\else
  \providecommand{\doi}{doi: \begingroup \urlstyle{rm}\Url}\fi

\bibitem[Ackermann et~al.(2025)Ackermann, Kulhanek, Cai, Haofei, Pollefeys,
  Wetzstein, Guibas, and Peng]{ackermann2025clsplats}
Jan Ackermann, Jonas Kulhanek, Shengqu Cai, Xu Haofei, Marc Pollefeys, Gordon
  Wetzstein, Leonidas Guibas, and Songyou Peng.
\newblock {CL-Splats}: Continual learning of {Gaussian} splatting with local
  optimization.
\newblock In \emph{Int. Conf. Comput. Vis.}, 2025.

\bibitem[Adelson and Bergen(1991)]{adelsonPlenopticFunctionElements1991}
Edward~H. Adelson and James~R. Bergen.
\newblock The plenoptic function and the elements of early vision.
\newblock In \emph{Computational Models of Visual Processing}. The MIT Press,
  1991.

\bibitem[Alcantarilla et~al.(2018)Alcantarilla, Stent, Ros, Arroyo, and
  Gherardi]{alcantarilla2018street}
Pablo~F Alcantarilla, Simon Stent, German Ros, Roberto Arroyo, and Riccardo
  Gherardi.
\newblock Street-view change detection with deconvolutional networks.
\newblock \emph{Auton. Robots}, 42\penalty0 (7):\penalty0 1301--1322, 2018.

\bibitem[Alpherts et~al.(2025)Alpherts, Ghebreab, and van
  Noord]{alpherts2025emplace}
Tim Alpherts, Sennay Ghebreab, and Nanne van Noord.
\newblock {EMPLACE}: Self-supervised urban scene change detection.
\newblock \emph{AAAI}, 39\penalty0 (2):\penalty0 1737--1745, 2025.

\bibitem[Brodersen et~al.(2010)Brodersen, Ong, Stephan, and
  Buhmann]{brodersen2010balanced}
Kay~Henning Brodersen, Cheng~Soon Ong, Klaas~Enno Stephan, and Joachim~M.
  Buhmann.
\newblock The balanced accuracy and its posterior distribution.
\newblock In \emph{Int. Conf. Pattern Recog.}, pages 3121--3124. IEEE, 2010.

\bibitem[Brown et~al.(2005)Brown, Szeliski, and Winder]{image_matching_2005}
M. Brown, R. Szeliski, and S. Winder.
\newblock Multi-image matching using multi-scale oriented patches.
\newblock In \emph{IEEE Conf. Comput. Vis. Pattern Recog.}, pages 510--517 vol.
  1, 2005.

\bibitem[Canny(1986)]{canny_1986}
John Canny.
\newblock A computational approach to edge detection.
\newblock \emph{IEEE Trans. Pattern Anal. Mach. Intell.}, PAMI-8\penalty0
  (6):\penalty0 679--698, 1986.

\bibitem[Caye~Daudt et~al.(2018)Caye~Daudt, Le~Saux, and
  Boulch]{caye_daudt_fully_2018}
Rodrigo Caye~Daudt, Bertr Le~Saux, and Alexandre Boulch.
\newblock Fully convolutional siamese networks for change detection.
\newblock In \emph{IEEE Int. Conf. Image Process.}, pages 4063--4067, Athens,
  2018. IEEE.

\bibitem[Chen et~al.(2021)Chen, Yang, and Stiefelhagen]{chen2021dr}
Shuo Chen, Kailun Yang, and Rainer Stiefelhagen.
\newblock Dr-tanet: Dynamic receptive temporal attention network for street
  scene change detection.
\newblock In \emph{IEEE Intell. Veh. Symp.}, pages 502--509. IEEE, 2021.

\bibitem[Cho et~al.(2025)Cho, Kim, and Kim]{cho2025zero}
Kyusik Cho, Dong~Yeop Kim, and Euntai Kim.
\newblock Zero-shot scene change detection.
\newblock \emph{AAAI}, 39\penalty0 (3):\penalty0 2509--2517, 2025.

\bibitem[Dansereau and Bruton(2004)]{dansereau_depth_2004}
Donald~G. Dansereau and L. Bruton.
\newblock Gradient-based depth estimation from {4D} light fields.
\newblock In \emph{IEEE Int. Symp. Circuit Sys.}, pages III--549, 2004.

\bibitem[Dansereau et~al.(2011)Dansereau, Mahon, Pizarro, and
  Williams]{dansereau2011plenoptic}
Donald~G. Dansereau, Ian Mahon, Oscar Pizarro, and Stefan~B. Williams.
\newblock Plenoptic flow: Closed-form visual odometry for light field cameras.
\newblock In \emph{IEEE/RSJ Int. Conf. Intell. Robot. Syst.}, pages 4455--4462.
  IEEE, 2011.

\bibitem[Dansereau et~al.(2016)Dansereau, Williams, and
  Corke]{dansereauSimpleChangeDetection2016}
Donald~G. Dansereau, Stefan~B. Williams, and Peter~I. Corke.
\newblock Simple change detection from mobile light field cameras.
\newblock \emph{Comput. Vis. Image Underst.}, 145:\penalty0 160--171, 2016.

\bibitem[Di~Zenzo(1986)]{dizenzo1986note}
Silvano Di~Zenzo.
\newblock A note on the gradient of a multi-image.
\newblock \emph{Comput. Vis. Graph. Image Process.}, 33\penalty0 (1):\penalty0
  116--125, 1986.

\bibitem[Eldar et~al.(1997)Eldar, Lindenbaum, Porat, and Zeevi]{eldar1997fps}
Y. Eldar, M. Lindenbaum, M. Porat, and Y.Y. Zeevi.
\newblock The farthest point strategy for progressive image sampling.
\newblock \emph{IEEE Trans. Image Process.}, 6\penalty0 (9):\penalty0
  1305--1315, 1997.

\bibitem[Freitas et~al.(2024)Freitas, De~Queiroz, Tabus, and
  Guillemot]{freitasComparativeAssessmentImplicit2024}
Davi~R. Freitas, Ricardo~L. De~Queiroz, Ioan Tabus, and Christine Guillemot.
\newblock A comparative assessment of implicit and explicit plenoptic scene
  representations.
\newblock In \emph{IEEE Int. Worksh. Multi. Sig. Proc.}, pages 1--6, West
  Lafayette, IN, USA, 2024. IEEE.

\bibitem[Freitas et~al.(2026)Freitas, Tabus, and Guillemot]{freitas_2026}
Davi~R. Freitas, Ioan Tabus, and Christine Guillemot.
\newblock Visibility-based geometry pruning of neural plenoptic scene
  representations.
\newblock \emph{IEEE Trans. Multimedia}, 28:\penalty0 1--16, 2026.

\bibitem[Fridovich-Keil et~al.(2022)Fridovich-Keil, Yu, Tancik, Chen, Recht,
  and Kanazawa]{fridovich_2022_plenoxels}
Sara Fridovich-Keil, Alex Yu, Matthew Tancik, Qinhong Chen, Benjamin Recht, and
  Angjoo Kanazawa.
\newblock Plenoxels: Radiance fields without neural networks.
\newblock In \emph{IEEE Conf. Comput. Vis. Pattern Recog.}, pages 5491--5500,
  2022.

\bibitem[Friedlander et~al.(2026)Friedlander, Shamir, and
  Fried]{friedlander2026goldilocs}
Almog Friedlander, Ariel Shamir, and Ohad Fried.
\newblock {GOLDILOCS}: General object-level detection and labeling of changes
  in scenes.
\newblock In \emph{Int. Conf. Learn. Represent.}, 2026.

\bibitem[Galappaththige et~al.(2025)Galappaththige, Lai, Windrim, Dansereau,
  S{\"u}nderhauf, and Miller]{Galappaththige_2025_CVPR}
Chamuditha~Jayanga Galappaththige, Jason Lai, Lloyd Windrim, Donald Dansereau,
  Niko S{\"u}nderhauf, and Dimity Miller.
\newblock Multi-view pose-agnostic change localization with zero labels.
\newblock In \emph{IEEE Conf. Comput. Vis. Pattern Recog.}, pages 11600--11610,
  2025.

\bibitem[Galappaththige et~al.(2026{\natexlab{a}})Galappaththige, Gottwald,
  Stehr, Heinert, Suenderhauf, Miller, and
  Rottmann]{galappaththige2026predictive}
Chamuditha~Jayanga Galappaththige, Thomas Gottwald, Peter Stehr, Edgar Heinert,
  Niko Suenderhauf, Dimity Miller, and Matthias Rottmann.
\newblock Predictive photometric uncertainty in gaussian splatting for novel
  view synthesis.
\newblock In \emph{Eur. Conf. Comput. Vis.}, 2026{\natexlab{a}}.

\bibitem[Galappaththige et~al.(2026{\natexlab{b}})Galappaththige, Lai, Patten,
  Dansereau, Suenderhauf, and Miller]{galappaththige2026gsdiff}
Chamuditha~Jayanga Galappaththige, Jason Lai, Timothy Patten, Donald~G.
  Dansereau, Niko Suenderhauf, and Dimity Miller.
\newblock From pixels to primitives: Scene change detection in {3D} {Gaussian}
  splatting.
\newblock \emph{arXiv preprint arXiv:2605.07203}, 2026{\natexlab{b}}.

\bibitem[Galappaththige et~al.(2026{\natexlab{c}})Galappaththige, Lai, Windrim,
  Dansereau, Suenderhauf, and Miller]{galappaththige2025changes}
Chamuditha~Jayanga Galappaththige, Jason Lai, Lloyd Windrim, Donald~G.
  Dansereau, Niko Suenderhauf, and Dimity Miller.
\newblock Changes in real time: Online scene change detection with multi-view
  fusion.
\newblock In \emph{IEEE Conf. Comput. Vis. Pattern Recog.}, pages 32246--32256,
  2026{\natexlab{c}}.

\bibitem[Gil et~al.(2010)Gil, Mozos, Ballesta, and Reinoso]{gil2010comparative}
Arturo Gil, Oscar~Martinez Mozos, Monica Ballesta, and Oscar Reinoso.
\newblock A comparative evaluation of interest point detectors and local
  descriptors for visual {SLAM}.
\newblock \emph{Mach. Vis. Appl.}, 21\penalty0 (6):\penalty0 905--920, 2010.

\bibitem[Gulrajani and Lopez-Paz(2021)]{gulrajani2021in}
Ishaan Gulrajani and David Lopez-Paz.
\newblock In search of lost domain generalization.
\newblock In \emph{Int. Conf. Learn. Represent.}, 2021.

\bibitem[Harris and Stephens(1988)]{harris1988combined}
Christopher~G. Harris and M.~J. Stephens.
\newblock A combined corner and edge detector.
\newblock In \emph{Alvey Vision Conference}, 1988.

\bibitem[Hattori-Nagao et~al.(2026)Hattori-Nagao, Oda, Eguchi, Nagao, and
  Kakuta]{hattori_2026}
Satoko Hattori-Nagao, Kazuo Oda, Tomoaki Eguchi, Takanobu Nagao, and Satomi
  Kakuta.
\newblock Render-to-real image-based change detection of outdoor infrastructure
  using {3D} {Gaussian} splatting.
\newblock \emph{Int. Arch. Photogramm. Remote Sens. Spatial Inf. Sci.},
  XLIX-B2-2026:\penalty0 771--777, 2026.

\bibitem[Huang et~al.(2023)Huang, Jiang, Zhao, Wang, Zhang, and
  Guo]{huangCNERFRepresentingScene2023}
Rui Huang, Binbin Jiang, Qingyi Zhao, William Wang, Yuxiang Zhang, and Qing
  Guo.
\newblock C-{NeRF}: Representing scene changes as directional consistency
  difference-based {NeRF}.
\newblock \emph{arXiv preprint arXiv:2312.02751}, 2023.

\bibitem[Jeong and Lee(2006)]{jeong_2006}
Woo~Yeon Jeong and Kyoung~Mu Lee.
\newblock Visual {SLAM} with line and corner features.
\newblock In \emph{IEEE/RSJ Int. Conf. Intell. Robot. Syst.}, pages 2570--2575,
  2006.

\bibitem[Jiang et~al.(2025)Jiang, Huang, Zhao, and
  Zhang]{jiangGaussianDifferenceFind2025}
Binbin Jiang, Rui Huang, Qingyi Zhao, and Yuxiang Zhang.
\newblock Gaussian difference: Find any change instance in {3D} scenes.
\newblock In \emph{ICASSP}, pages 1--5, Kothaguda, India, 2025. IEEE.

\bibitem[Johannsen et~al.(2015)Johannsen, Sulc, and Goldluecke]{johannsen_2015}
Ole Johannsen, Antonin Sulc, and Bastian Goldluecke.
\newblock On linear structure from motion for light field cameras.
\newblock In \emph{Int. Conf. Comput. Vis.}, pages 720--728, 2015.

\bibitem[Kannan and Min(2025)]{kannan_2025}
Shyam~Sundar Kannan and Byung-Cheol Min.
\newblock {ZeroSCD}: Zero-shot street scene change detection.
\newblock In \emph{IEEE Int. Conf. Rob. Autom.}, pages 4665--4671, 2025.

\bibitem[Kerbl et~al.(2023)Kerbl, Kopanas, Leimkuehler, and
  Drettakis]{kerbl3DGaussianSplatting2023}
Bernhard Kerbl, Georgios Kopanas, Thomas Leimkuehler, and George Drettakis.
\newblock {3D Gaussian} splatting for real-time radiance field rendering.
\newblock \emph{ACM Trans. Graph.}, 42\penalty0 (4):\penalty0 1--14, 2023.

\bibitem[Kheradmand et~al.(2024)Kheradmand, Rebain, Sharma, Sun, Tseng, Isack,
  Kar, Tagliasacchi, and Yi]{kheradmand20243d}
Shakiba Kheradmand, Daniel Rebain, Gopal Sharma, Weiwei Sun, Yang-Che Tseng,
  Hossam Isack, Abhishek Kar, Andrea Tagliasacchi, and Kwang~Moo Yi.
\newblock {3D Gaussian} splatting as markov chain monte carlo.
\newblock In \emph{Adv. Neural Inform. Process. Syst.}, 2024.
\newblock Spotlight Presentation.

\bibitem[Kim and Kim(2025)]{Kim_2025_CVPR}
Jae-Woo Kim and Ue-Hwan Kim.
\newblock Towards generalizable scene change detection.
\newblock In \emph{IEEE Conf. Comput. Vis. Pattern Recog.}, pages 24463--24473,
  2025.

\bibitem[Kirillov et~al.(2023)Kirillov, Mintun, Ravi, Mao, Rolland, Gustafson,
  Xiao, Whitehead, Berg, Lo, Dollar, and Girshick]{kirillovSegmentAnything2023}
Alexander Kirillov, Eric Mintun, Nikhila Ravi, Hanzi Mao, Chloe Rolland, Laura
  Gustafson, Tete Xiao, Spencer Whitehead, Alexander~C. Berg, Wan-Yen Lo, Piotr
  Dollar, and Ross Girshick.
\newblock Segment anything.
\newblock In \emph{Int. Conf. Comput. Vis.}, pages 4015--4026, 2023.

\bibitem[Kruse et~al.(2024)Kruse, Rudolph, Woiwode, and
  Rosenhahn]{kruseSplatPoseDetectPoseAgnostic}
Mathis Kruse, Marco Rudolph, Dominik Woiwode, and Bodo Rosenhahn.
\newblock {SplatPose} \& detect: Pose-agnostic {3D} anomaly detection.
\newblock In \emph{IEEE Conf. Comput. Vis. Pattern Recog. Worksh.}, pages
  3950--3960, 2024.

\bibitem[Laptev(2005)]{laptev2005space}
Ivan Laptev.
\newblock On space-time interest points.
\newblock \emph{Int. J. Comput. Vis.}, 64\penalty0 (2):\penalty0 107--123,
  2005.

\bibitem[Lee and Kim(2024)]{lee2024semi}
Seonhoon Lee and Jong-Hwan Kim.
\newblock Semi-supervised scene change detection by distillation from
  feature-metric alignment.
\newblock In \emph{IEEE Winter Conf. App. Comput. Vis.}, pages 1226--1235,
  2024.

\bibitem[Lei et~al.(2020)Lei, Peng, Zhang, Ke, and Li]{lei2020hierarchical}
Yinjie Lei, Duo Peng, Pingping Zhang, Qiuhong Ke, and Haifeng Li.
\newblock Hierarchical paired channel fusion network for street scene change
  detection.
\newblock \emph{IEEE Trans. Image Process.}, 30:\penalty0 55--67, 2020.

\bibitem[Leroy et~al.(2024)Leroy, Cabon, and Revaud]{mast3r}
Vincent Leroy, Yohann Cabon, and Jerome Revaud.
\newblock Grounding image matching in {3D} with {MASt3R}.
\newblock In \emph{Eur. Conf. Comput. Vis.}, page 71–91, Berlin, Heidelberg,
  2024. Springer-Verlag.

\bibitem[Levoy and Hanrahan(1996)]{levoyLightFieldRendering1996}
Marc Levoy and Pat Hanrahan.
\newblock Light field rendering.
\newblock In \emph{ACM SIGGRAPH}, pages 31--42, New Orleans, USA, 1996.
  Association for Computing Machinery.

\bibitem[Li et~al.(2025)Li, Wu, Meng, Gao, Zhang, Wang, and
  Zhang]{li2025instancegaussian}
Haijie Li, Yanmin Wu, Jiarui Meng, Qiankun Gao, Zhiyao Zhang, Ronggang Wang,
  and Jian Zhang.
\newblock Instancegaussian: Appearance-semantic joint {Gaussian} representation
  for {3D} instance-level perception.
\newblock In \emph{IEEE Conf. Comput. Vis. Pattern Recog.}, 2025.

\bibitem[Lin et~al.(2025)Lin, Garg, Chin, and Dayoub]{linRobustSceneChange2024}
Chun-Jung Lin, Sourav Garg, Tat-Jun Chin, and Feras Dayoub.
\newblock Robust scene change detection using visual foundation models and
  cross-attention mechanisms.
\newblock In \emph{IEEE Int. Conf. Rob. Autom.}, pages 8337--8343, Atlanta,
  USA, 2025. IEEE.

\bibitem[Liu et~al.(2024)Liu, Hu, Chen, and
  Zelek]{liuSplatPoseRealtimeImageBased2024}
Yizhe Liu, Yan~Song Hu, Yuhao Chen, and John Zelek.
\newblock {SplatPose+}: Real-time image-based pose-agnostic {3D} anomaly
  detection.
\newblock In \emph{Eur. Conf. Comput. Vis. Worksh.}, pages 378--391, 2024.

\bibitem[Liu et~al.(2025)Liu, Chen, Gao, Yang, and Zheng]{liu2025leveraging}
Ziling Liu, Ziwei Chen, Mingqi Gao, Jinyu Yang, and Feng Zheng.
\newblock Leveraging geometric priors for unaligned scene change detection.
\newblock \emph{arXiv preprint arXiv:2509.11292}, 2025.

\bibitem[Lu et~al.(2025)Lu, Ye, and Leonard]{lu3DGSCD3DGaussian2025}
Ziqi Lu, Jianbo Ye, and John Leonard.
\newblock {3DGS-CD}: {3D Gaussian} splatting-based change detection for
  physical object rearrangement.
\newblock \emph{IEEE Robot. Autom. Letters}, 10\penalty0 (3):\penalty0
  2662--2669, 2025.

\bibitem[Ma et~al.(2021)Ma, Jiang, Fan, Jiang, and Yan]{ma2021image}
Jiayi Ma, Xingyu Jiang, Aoxiang Fan, Junjun Jiang, and Junchi Yan.
\newblock Image matching from handcrafted to deep features: A survey.
\newblock \emph{Int. J. Comput. Vis.}, 129\penalty0 (1):\penalty0 23--79, 2021.

\bibitem[Ma et~al.(2018)Ma, Smith, and Gupta]{Ma_2018_ECCV}
Sizhuo Ma, Brandon~M. Smith, and Mohit Gupta.
\newblock {3D} scene flow from {4D} light field gradients.
\newblock In \emph{Eur. Conf. Comput. Vis.}, 2018.

\bibitem[Martinson and Lauren(2024)]{martinsonMeaningfulChangeDetection2024}
E. Martinson and P. Lauren.
\newblock Meaningful {Change Detection} in indoor environments using {CLIP}
  models and {NeRF}-based image synthesis.
\newblock In \emph{Int. Conf. Ubiq. Rob.}, pages 603--610, 2024.

\bibitem[Matsuki et~al.(2024)Matsuki, Murai, Kelly, and
  Davison]{matsuki2024gaussian}
Hidenobu Matsuki, Riku Murai, Paul~HJ Kelly, and Andrew~J Davison.
\newblock Gaussian splatting {SLAM}.
\newblock In \emph{IEEE Conf. Comput. Vis. Pattern Recog.}, pages 18039--18048,
  2024.

\bibitem[Mildenhall et~al.(2020)Mildenhall, Srinivasan, Tancik, Barron,
  Ramamoorthi, and Ng]{mildenhallNeRFRepresentingScenes2020}
Ben Mildenhall, Pratul~P. Srinivasan, Matthew Tancik, Jonathan~T. Barron, Ravi
  Ramamoorthi, and Ren Ng.
\newblock {{NeRF}}: Representing scenes as neural radiance fields for view
  synthesis.
\newblock In \emph{Eur. Conf. Comput. Vis.}, pages 405--421, Cham, Switzerland,
  2020. Springer International Publishing.

\bibitem[Moenne-Loccoz et~al.(2024)Moenne-Loccoz, Mirzaei, Perel, de~Lutio,
  Martinez~Esturo, State, Fidler, Sharp, and Gojcic]{moenne20243d}
Nicolas Moenne-Loccoz, Ashkan Mirzaei, Or Perel, Riccardo de Lutio, Janick
  Martinez~Esturo, Gavriel State, Sanja Fidler, Nicholas Sharp, and Zan Gojcic.
\newblock {3D Gaussian} ray tracing: Fast tracing of particle scenes.
\newblock \emph{ACM Trans. Graph.}, 43\penalty0 (6):\penalty0 1--19, 2024.

\bibitem[Naylor et~al.(2026)Naylor, Ila, and Dansereau]{naylor2025surf}
Jack Naylor, Viorela Ila, and Donald~G. Dansereau.
\newblock {Surf-NeRF}: Surface regularised neural radiance fields.
\newblock \emph{IEEE Conf. Comput. Vis. Pattern Recog. Worksh.}, pages
  258--268, 2026.

\bibitem[Neumann et~al.(2002)Neumann, Fermuller, and Aloimonos]{neumann_2002}
J. Neumann, C. Fermuller, and Y. Aloimonos.
\newblock A hierarchy of cameras for {3D} photography.
\newblock In \emph{Int. Symp. {3D} Data Process., Vis. Transm.}, pages 2--11,
  2002.

\bibitem[Oquab et~al.(2024)Oquab, Darcet, Moutakanni, Vo, Szafraniec, Khalidov,
  Fernandez, Haziza, Massa, El-Nouby, et~al.]{oquab2024dinov2}
Maxime Oquab, Timoth{\'e}e Darcet, Th{\'e}o Moutakanni, Huy Vo, Marc
  Szafraniec, Vasil Khalidov, Pierre Fernandez, Daniel Haziza, Francisco Massa,
  Alaaeldin El-Nouby, et~al.
\newblock {DINOv2}: Learning robust visual features without supervision.
\newblock \emph{Trans. Mach. Learn. Res.}, 2024.

\bibitem[Park et~al.(2021)Park, Jang, Yoo, Lee, Kim, and
  Kim]{park2021changesim}
Jin-Man Park, Jae-hyuk Jang, Sahng-Min Yoo, Sun-Kyung Lee, Ue-hwan Kim, and
  Jong-Hwan Kim.
\newblock {ChangeSim}: Towards end-to-end online scene change detection in
  industrial indoor environments.
\newblock In \emph{IEEE/RSJ Int. Conf. Intell. Robot. Syst.} IEEE, 2021.

\bibitem[Petrovska and Jutzi(2025)]{petrovska_2025}
I. Petrovska and B. Jutzi.
\newblock {3D Gaussian} splatting methods for real-world scenarios.
\newblock \emph{ISPRS Ann. Photogramm. Remote Sens. Spatial Inf. Sci.},
  X-G-2025:\penalty0 641--648, 2025.

\bibitem[Radke et~al.(2005)Radke, Andra, Al-Kofahi, and Roysam]{radke2005image}
Richard~J Radke, Srinivas Andra, Omar Al-Kofahi, and Badrinath Roysam.
\newblock Image change detection algorithms: A systematic survey.
\newblock \emph{IEEE Trans. Image Process.}, 14\penalty0 (3):\penalty0
  294--307, 2005.

\bibitem[Sachdeva and Zisserman(2023{\natexlab{a}})]{sachdeva2023change}
Ragav Sachdeva and Andrew Zisserman.
\newblock The change you want to see.
\newblock In \emph{IEEE Winter Conf. App. Comput. Vis.}, pages 3993--4002,
  2023{\natexlab{a}}.

\bibitem[Sachdeva and Zisserman(2023{\natexlab{b}})]{sachdevaChangeYouWant2023}
Ragav Sachdeva and Andrew Zisserman.
\newblock The change you want to see (now in {3D}).
\newblock In \emph{Int. Conf. Comput. Vis. Worksh.}, pages 2052--2061, Paris,
  France, 2023{\natexlab{b}}. IEEE.

\bibitem[Sakurada and Okatani(2015)]{sakurada_change_2015}
Ken Sakurada and Takayuki Okatani.
\newblock Change detection from a street image pair using {CNN} features and
  superpixel segmentation.
\newblock In \emph{Brit. Mach. Vis. Conf.}, pages 61.1--61.12, Swansea, 2015.
  British Machine Vision Association.

\bibitem[Sakurada et~al.(2020)Sakurada, Shibuya, and
  Wang]{sakuradaWeaklySupervisedSilhouettebased2020}
Ken Sakurada, Mikiya Shibuya, and Weimin Wang.
\newblock Weakly supervised silhouette-based semantic scene change detection.
\newblock In \emph{IEEE Int. Conf. Rob. Autom.}, pages 6861--6867, Paris,
  France, 2020. IEEE.

\bibitem[Sch{\"o}nberger and Frahm(2016)]{schonberger_sfm_2016}
Johannes~L. Sch{\"o}nberger and Jan-Michael Frahm.
\newblock Structure-from-motion revisited.
\newblock In \emph{IEEE Conf. Comput. Vis. Pattern Recog.}, pages 4104--4113,
  2016.

\bibitem[Shi and {Tomasi}(1994)]{shiGoodFeaturesTrack1994}
Jianbo Shi and {Tomasi}.
\newblock Good features to track.
\newblock In \emph{IEEE Conf. Comput. Vis. Pattern Recog.}, pages 593--600,
  1994.

\bibitem[Szeliski and Cohen(1996)]{szeliskiLumigraph1996}
Richard Szeliski and Michael~F Cohen.
\newblock The lumigraph.
\newblock In \emph{ACM SIGGRAPH}, pages 43--54, New Orleans, USA, 1996.
  Association for Computing Machinery.

\bibitem[Taneja et~al.(2011)Taneja, Ballan, and Pollefeys]{taneja2011image}
Aparna Taneja, Luca Ballan, and Marc Pollefeys.
\newblock Image based detection of geometric changes in urban environments.
\newblock In \emph{2011 International Conference on Computer Vision}, pages
  2336--2343. IEEE, 2011.

\bibitem[Tsai et~al.(2019)Tsai, Dansereau, Peynot, and Corke]{tsai_2019}
Dorian Tsai, Donald~G. Dansereau, Thierry Peynot, and Peter Corke.
\newblock Distinguishing refracted features using light field cameras with
  application to structure from motion.
\newblock \emph{IEEE Robot. Autom. Letters}, 4\penalty0 (2):\penalty0 177--184,
  2019.

\bibitem[Varghese et~al.(2018{\natexlab{a}})Varghese, Gubbi, Ramaswamy, and
  Balamuralidhar]{varghese2018changenet}
Ashley Varghese, Jayavardhana Gubbi, Akshaya Ramaswamy, and P Balamuralidhar.
\newblock {ChangeNet}: A deep learning architecture for visual change
  detection.
\newblock In \emph{Eur. Conf. Comput. Vis. Worksh.}, pages 0--0,
  2018{\natexlab{a}}.

\bibitem[Varghese et~al.(2018{\natexlab{b}})Varghese, Gubbi, Ramaswamy, and
  Balamuralidhar]{vargheseChangeNetDeepLearning2018}
Ashley Varghese, Jayavardhana Gubbi, Akshaya Ramaswamy, and P. Balamuralidhar.
\newblock {ChangeNet}: A deep learning architecture for visual change
  detection.
\newblock In \emph{Eur. Conf. Comput. Vis. Worksh.}, pages 129--145. Springer
  International Publishing, Munich, Germany, 2018{\natexlab{b}}.

\bibitem[Wang et~al.(2023)Wang, Gao, and Wang]{wangHowReduceChange2023}
Guo-Hua Wang, Bin-Bin Gao, and Chengjie Wang.
\newblock How to reduce change detection to semantic segmentation.
\newblock \emph{Pattern Recognition}, 138:\penalty0 109384, 2023.

\bibitem[Wang et~al.(2025)Wang, Chen, Karaev, Vedaldi, Rupprecht, and
  Novotny]{wang2025vggt}
Jianyuan Wang, Minghao Chen, Nikita Karaev, Andrea Vedaldi, Christian
  Rupprecht, and David Novotny.
\newblock Vggt: Visual geometry grounded transformer.
\newblock In \emph{IEEE Conf. Comput. Vis. Pattern Recog.}, 2025.

\bibitem[Wang et~al.(2026)Wang, Zhou, Zhu, Chang, Zhou, Li, Chen, Pang, Shen,
  and He]{wang2026pi}
Yifan Wang, Jianjun Zhou, Haoyi Zhu, Wenzheng Chang, Yang Zhou, Zizun Li, Junyi
  Chen, Jiangmiao Pang, Chunhua Shen, and Tong He.
\newblock \${\textbackslash}pi{\textasciicircum}3\$: Permutation-equivariant
  visual geometry learning.
\newblock In \emph{Int. Conf. Learn. Represent.}, 2026.

\bibitem[Wu et~al.(2025)Wu, Esturo, Mirzaei, Moenne-Loccoz, and
  Gojcic]{wu20253dgut}
Qi Wu, Janick~Martinez Esturo, Ashkan Mirzaei, Nicolas Moenne-Loccoz, and Zan
  Gojcic.
\newblock {3DGUT}: Enabling distorted cameras and secondary rays in {Gaussian}
  splatting.
\newblock In \emph{IEEE Conf. Comput. Vis. Pattern Recog.}, pages 26036--26046,
  2025.

\bibitem[Wu et~al.(2026)Wu, Lin, Che, Tiwari, Zou, Wang, and
  Hoiem]{wu2026scenediff}
Yuqun Wu, Chih-hao Lin, Henry Che, Aditi Tiwari, Chuhang Zou, Shenlong Wang,
  and Derek Hoiem.
\newblock {SceneDiff}: A benchmark and method for multiview object change
  detection.
\newblock In \emph{Eur. Conf. Comput. Vis.}, 2026.

\bibitem[Ye et~al.(2025)Ye, Li, Kerr, Turkulainen, Yi, Pan, Seiskari, Ye, Hu,
  Tancik, et~al.]{ye2025gsplat}
Vickie Ye, Ruilong Li, Justin Kerr, Matias Turkulainen, Brent Yi, Zhuoyang Pan,
  Otto Seiskari, Jianbo Ye, Jeffrey Hu, Matthew Tancik, et~al.
\newblock gsplat: An open-source library for {Gaussian} splatting.
\newblock \emph{J. Mach. Learn. Res.}, 26\penalty0 (34):\penalty0 1--17, 2025.

\bibitem[Zhou et~al.(2023)Zhou, Li, Jiang, Wang, Zhou, Zhang, and
  Zhao]{zhouPADDatasetBenchmark2023}
Qiang Zhou, Weize Li, Lihan Jiang, Guoliang Wang, Guyue Zhou, Shanghang Zhang,
  and Hao Zhao.
\newblock {{PAD}}: A dataset and benchmark for pose-agnostic anomaly detection.
\newblock In \emph{Adv. Neural Inform. Process. Syst.}, pages 44558--44571, New
  Orleans, USA, 2023.

\bibitem[Zhou et~al.(2025)Zhou, Ni, Zhang, Chen, and Huang]{zhou20253d}
Zirui Zhou, Junfeng Ni, Shujie Zhang, Yixin Chen, and Siyuan Huang.
\newblock {3D} scene change modeling with consistent multi-view aggregation.
\newblock In \emph{Int. Conf. 3D Vis.}, 2025.

\end{thebibliography}
